%% file: iclr2026_conference.tex
\documentclass{article} 
\usepackage{iclr2026_conference,times}

\input{math_commands.tex}

\usepackage{wrapfig}
\usepackage{hyperref}
\usepackage{url}
\usepackage{diagbox}
\usepackage{multirow}
\usepackage{amsmath,amssymb}
\usepackage{subcaption}
\usepackage{graphicx}
\usepackage{arydshln}
\usepackage{xcolor}
\usepackage{soul}
\hypersetup{colorlinks=true, linkcolor=black, citecolor=blue, urlcolor=blue}
\usepackage{caption}
\usepackage[ruled,vlined]{algorithm2e}

\title{Out-of-distribution transfer of PDE foundation models to material dynamics under extreme loading}

\author{
Mahindra Rautela\textsuperscript{*1},
Alexander Most $^{2}$, 
Siddharth Mansingh $^{2}$,
Aleksandra Pachalieva $^{3}$, \\
\textbf{Bradley Love}$^{2}$,
\textbf{Daniel O’Malley} $^{3}$, 
\textbf{Alexander Scheinker}$^{1}$, 
\textbf{Kyle Hickmann} $^{4}$, \\
\textbf{Diane Oyen}$^{2}$,
\textbf{Nathan Debardeleben}$^{5}$,
\textbf{Earl Lawrence}$^{2}$
\textbf{Ayan Biswas}$^{2}$ \\
$^1$AOT-IC Group, $^2$CAI Division, $^3$EES Divison, $^4$XCP Division, $^5$HPC Division \\
Los Alamos National Laboratory, Los Alamos, New Mexico, US, 87545 \\
{\tt\small\{mrautela,amost,smansingh,apachalieva,love,omalled,ascheink,} \\
\tt \small hickmank,doyen,ndebard,earl,ayan\}@lanl.gov}

\iclrfinalcopy 

\begin{document}

\maketitle

\begin{abstract}
Most PDE foundation models are pretrained and fine-tuned on fluid-centric benchmarks. Their utility under extreme-loading material dynamics remains unclear. We benchmark out-of-distribution transfer on two discontinuity-dominated regimes in which shocks, evolving interfaces, and fracture produce highly non-smooth fields: shock-driven multi-material interface dynamics (perturbed layered interface or PLI) and dynamic fracture/failure evolution (FRAC). We formulate the downstream task as terminal-state prediction, i.e., learning a long-horizon map that predicts the final state directly from the first snapshot without intermediate supervision. Using a unified training and evaluation protocol, we evaluate two open-source pretrained PDE foundation models, POSEIDON and MORPH, and compare fine-tuning from pretrained weights against training from scratch across training-set sizes to quantify sample efficiency under distribution shift.
\end{abstract}

\section{Introduction}
\label{sec:intro}

\emph{The physics of material dynamics under extreme loading is challenging.}
Many mission and safety-critical systems (e.g., spacecraft structures and thermal protection, hypersonic vehicles, and explosive or impact-driven experiments) operate in regimes where materials experience extreme transient loading, including shock and blast loading, rapid energy deposition, impact, and impulsive mechanical forcing \citep{kalthoff1986fracture, sterbentz2022design, demmie2007approach}. These systems are challenging for both classical simulation and learning-based surrogates because their solutions are dominated by shocks/steep gradients, evolving multi-material interfaces and morphology changes. Accurate prediction of the terminal (late-time) state is central in such problems for inverse estimation, uncertainty quantification, and design optimization.

\emph{PDE foundation models aim to amortize operator learning via pretraining and fine-tuning.}
Standalone PDE surrogate models are widely used in scientific machine learning \citep{RAISSI2019686,chen2018neural,li2020fourier,khodkar2021data,goswami2022deep,oommen2022learning, maulik2021reduced, rautela2024conditional, kontolati2024learning, rautela2025time}. However, these surrogates are typically specialized to a particular PDE (or parameterized family), geometry, and discretization, and often require retraining or substantial re-tuning when the physics, resolution, or observational modality changes. This limitation motivates pretraining-based paradigms that amortize learning across PDE families. PDE foundation models (PDE-FMs) are task-agnostic surrogates pretrained on large, diverse PDE solution corpora to learn transferable representations, then adapted via fine-tuning to downstream tasks \citep{subramanian2023towards}. This improves sample efficiency and generalization, especially when transferring to previously unseen PDE families, parameter regimes, or geometries \citep{hao2024dpot,herde2024poseidon,rautela2025morph}. Despite this promise, evaluation of PDE foundation models under out-of-distribution datasets is still limited.

\emph{Progress on PDE-FMs depends on rigorous out-of-distribution evaluation.}
PDE foundation models are often presented as broadly transferable surrogates, but their practical usefulness hinges on robustness under distribution shift across PDE families, discretizations, geometries, boundary conditions, and sensing modalities. While pretraining can learn transferable representations, it can also imprint inductive biases from the pretraining distribution, leading to brittle behavior under shifts in physics, resolution, or observation type. In the LLM literature, model utility is established through standardized multiple downstream tasks spanning coding \citep{chen2021evaluating}, broad multitask knowledge \citep{hendrycks2020measuring}, mathematics \citep{hendrycks2021measuring}, and challenging reasoning tasks \citep{suzgun2023challenging}. Analogously, PDE foundation models should be assessed using diverse dataset suites that explicitly probe out-of-distribution transfer, stability, and failure modes before they are considered general-purpose surrogates.

\emph{Why benchmark OOD transfer on extreme-loading material dynamics?}
Despite rapid progress, most publicly available pretraining corpora and standardized evaluations for PDE surrogates remain dominated by \emph{fluid-centric} or \emph{smooth-field} regimes (e.g., advection--diffusion, Burgers, incompressible/compressible Navier--Stokes, and related canonical flows). Large collections such as The Well \citep{ohana2024well} and benchmark suites such as PDEBench \citep{takamoto2022pdebench} and PDEGym \citep{herde2024poseidon} have accelerated PDE-surrogate research, but they only partially capture the discontinuity- and interface-dominated dynamics that arise in multi-material shock physics and dynamic fracture/failure. Consequently, it remains unclear how well PDE foundation models pretrained on fluid-dominated distributions transfer to extreme-loading mechanics, where the dynamics are governed by (i) shocks and steep gradients, (ii) rate-dependent inelastic response at high strain rates, (iii) a large dynamic range of spatial and temporal scales, (iv) free-boundary evolution driven by material interfaces and cracks, (v) localization phenomena that control global outcomes, and (vi) strong sensitivity of late-time morphology to small early-time perturbations \citep{sterbentz2023linear,marcato2025foundation}. Moreover, whereas most downstream evaluations emphasize autoregressive rollouts, we instead focus on terminal-state prediction as a practically relevant long-horizon task, since it is often more consequential for design optimization and uncertainty quantification.

\emph{Problem setting and evaluation protocol.}
We fine-tune PDE foundation models for terminal-state prediction and formulate the downstream task as \emph{first-frame to final-frame} long-horizon operator learning. Given an initial multi-field state at an early time $t_0$, the model predicts the terminal state at $t_T$ without intermediate supervision, thereby learning a long-horizon mapping. In this work, we introduce a unified evaluation protocol for out-of-distribution transfer on extreme-loading material dynamics using two complementary open-source datasets, perturbed layered interface (PLI) and dynamic fracture/failure evolution (FRAC). Under this protocol, we benchmark two predominantly fluid-pretrained, open-source PDE foundation models (MORPH and POSEIDON) by comparing fine-tuning versus random initialization across varying training-set sizes, thereby quantifying sample efficiency and generalization.

\section{Methods}
\subsection{Datasets}
\subsubsection{Perturbed Layered Interface.}
To probe the extreme-loading regime, we use the Perturbed Layered Interface (PLI) dataset, a large-scale physics dataset of two-dimensional axisymmetric multi-material simulations designed to capture high-explosive driven shock propagation through complex targets \citep{HEAT}. Each sample consists of a spatiotemporal trajectory of multi-field state variables, including thermodynamic and kinematic fields such as density, pressure, temperature, and velocity, along with additional derived mechanics-related quantities. This structure makes PLI well suited for learning long-horizon mappings from early-time conditions to late-time (terminal) states. The trajectories exhibit key phenomena including momentum transfer, shock propagation, plastic deformation, and thermal effects. The underlying dynamics are strongly interface controlled, since shock and interface interactions can trigger Richtmyer-Meshkov-type instabilities and jetting. Across the dataset, geometry varies substantially while the material set is fixed, comprising copper, aluminum, stainless steel, a generic polymer, a generic high explosive, and an air background.

The dataset comprises 5,293 simulations, each represented by 38 channels over 100 time steps at a spatial resolution of 
1120 $\times$ 400. For terminal-state prediction, we use only the first and last frames of each trajectory. Unless otherwise noted, we use the `av-density' channel, which encodes the mixture-averaged density across materials. We partition the dataset into training, validation, and test splits using an 80/10/10 split. A representative spatiotemporal trajectory is shown in Fig.~\ref{fig:pli} (Appendix~\ref{app:dataset}), where columns correspond to equally spaced time steps and each row visualizes the density field of multiple materials.

\subsubsection{Dynamic fracture and failure evolution.}
Complementing PLI, we use the Material Fracturing and Failure Simulation Datasets (FRAC), which focus on fracture initiation, propagation, and interaction across multiple materials and loading conditions \citep{hill2025material}. FRAC is generated with two solver families that differ in formulation and fidelity: a phase-field fracture model and a combined finite-discrete element method (FDEM, HOSS), spanning structured (Cartesian) and unstructured representations. The dataset covers diverse material classes (including brittle and ductile responses) and boundary/loading configurations (e.g., uniaxial and biaxial tension in the phase-field subset), with randomized initial fracture patterns to drive diverse crack-network evolution. Because the late-time crack topology and damage localization are path-dependent and governed by moving fronts (crack tips) and evolving discontinuities, FRAC provides a stringent testbed for long-horizon prediction under extreme, highly nonlinear solid mechanics \citep{marcato2025foundation}.  

The dataset contains multi-material simulations. In this work, we focus on the tungsten subset under horizontal boundary condition, which comprises approximately 200K simulations with variable numbers of time steps across trajectories. We partition this subset into training, validation, and test splits using an 80/10/10 split. A representative spatiotemporal trajectory is shown in Fig.~\ref{fig:frac} (Appendix~\ref{app:dataset}), where we visualize three example simulations. Within each row, frames correspond to successive time steps, annotated with their characteristic-time.

\subsection{Models}
We study two pretrained PDE foundation models with comparable capacity: POSEIDON and MORPH. Both models are transformer-based operator learners pretrained on different PDE corpora and adapted to our extreme-loading regimes via supervised fine-tuning.

\subsubsection{POSEIDON PDE-FM} 
The model is built on the scalable Operator Transformer (scOT), a hierarchical multiscale vision transformer designed to approximate solution operators of time-dependent PDEs \citep{herde2024poseidon}. Given an initial condition `a' and a lead time `t', scOT produces an estimate of the state S(t,a). The input field is partitioned into patches and embedded into latent tokens, which are processed through multiple stages of SwinV2 blocks with shifted-window self-attention, together with multiscale token resolutions via patch merging/expansion. Time dependence is incorporated through lead-time conditioning in normalization. POSEIDON uses time-conditioned layer normalization whose affine parameters are functions of the queried lead time, enabling a single model to represent multiple prediction horizons. The pretraining procedure leverages the semigroup structure of PDE evolution through an all-to-all pairing strategy over time snapshots, exposing the model to diverse $\Delta t$ mappings within each trajectory.

POSEIDON is pretrained on six datasets from PDEGym: two incompressible Navier–Stokes variants (NS-Sines and NS-Gauss) and four compressible Euler variants (CE-RP, CE-KH, CE-CRP, and CE-Gauss). Pretraining uses roughly 20,000 Navier–Stokes trajectories and 10,000 Euler trajectories (about 80,000 trajectories in total), with inputs and outputs standardized to four fluid variables (density, u-velocity, v-velocity, and pressure) on a 128×128 grid. Downstream transfer is assessed by fine-tuning the pretrained model on 15 out-of-distribution PDEGym tasks spanning three settings: (i) new Navier–Stokes initial-condition families, (ii) modified fluid dynamics with additional physics (such as tracer transport, Kolmogorov forcing, or gravity), and (iii) new PDE classes, including wave propagation, Allen–Cahn reaction–diffusion, and steady elliptic problems such as airfoil flow, Poisson, and Helmholtz.

In our experiments, we initialize from the released \emph{POSEIDON-T ($\approx$21M parameters)} checkpoint and fine-tune it for terminal-state prediction by conditioning on the final lead time. We use the default patch-size of 4 $\times$ 4. The reference POSEIDON-T implementation expects inputs in a fixed tensor format of size $4 \times 128 \times 128$ (channels $\times$ height $\times$ width). To accommodate PLI, where each example is provided as a single-channel field at native resolution $1 \times 1120 \times 400$, we replicate the channel dimension to obtain $4 \times 1120 \times 400$. We find that this setting yields better results than the alternative options. We then downsample to $4 \times 128 \times 128$ using bilinear interpolation on-the-fly. During evaluation, we upsample the predicted terminal field back to the original $1120 \times 400$ grid and compute the test MSE at the native resolution. For FRAC, inputs are already $1 \times 128 \times 128$, therefore, we only replicate the single channel to four channels ($4 \times 128 \times 128$) and apply no spatial rescaling. With a $4 \times 4 $ patch size, this results in $32 \times 32 = 1{,}024$ input patches per batch. 

\subsubsection{MORPH PDE-FM} 
The model is a modality-agnostic PDE foundation model that explicitly targets heterogeneous scientific data (varying spatial dimensionality, resolutions, and mixed scalar/vector fields) \citep{rautela2025morph}. Architecturally, MORPH combines: (i) component-wise convolutions over the component dimension to inject local inductive bias and jointly process scalar/vector channels; (ii) field-wise multi-head cross-attention that performs content-based fusion across physical fields, producing a single fused representation and naturally accommodating a variable number of fields; and (iii) axial attention that factorizes spatiotemporal self-attention along individual spatial and temporal axes, reducing compute while retaining long-range expressivity. In addition, MORPH includes built-in low-rank adapter (LoRA) layers to support parameter-efficient fine-tuning.

MORPH is pretrained on six heterogeneous datasets from PDEBench and The Well: diffusion-reaction-2D, shallow-water-2D, CFD-1D, incompressible CFD-2D, CFD-3D, and magnetohydrodynamics-3D. These datasets span 1D to 3D modalities (up to 128×128×128), with resolutions ranging from 1D grids of length 1024 to 512×512 2D grids, and include both scalar and vector fields. Transfer is evaluated by fine-tuning on additional targets drawn from PDEBench (diffusion-reaction-1D, Burger-equation-1D, CFD-2D, and CFD3D-turbulence), The Well (gray-scott diffusion-reaction-2D and turbulent-gravity-cooling-3D), and PDEGym (FNS-KF-2D, CE-RM, NS-PwC, and NS-SL).

In our experiments, we initialize \emph{MORPH-Ti ($\approx$7M parameters)} from the released checkpoint and fine-tune it end-to-end for terminal-state prediction. MORPH is pretrained with an $8 \times 8$ patchification scheme. Because the architecture is designed to accommodate heterogeneous modalities and resolutions, we fine-tune on the native spatial grids of PLI and FRAC. To control the tokenization granularity during fine-tuning, we consider patch sizes of $8 \times 8$ and $4 \times 4$, which in our implementation correspond to approximately $3{,}500$ and $1{,}024$ patches per batch, respectively.

\section{Experiments}
\subsection{Settings}
We adopt an identical fine-tuning protocol for POSEIDON and MORPH. All models are optimized with AdamW (weight decay = $10^{-2}$) and a mean-squared error (MSE) objective. We employ a learning-rate schedule consisting of a 5-epoch linear warmup from $10^{-7}$ to the peak learning rate, followed by cosine decay for the remainder of training. The peak learning rates found suitable were $10^{-4}$ for PLI and $10^{-3}$ for FRAC. We additionally use early stopping based on validation loss with a patience of 10 epochs. All experiments use data-parallel training on a single node with 8 NVIDIA H100 GPUs. We use an effective batch size of 8 for PLI and 512 for FRAC for both models.

We consider two experimental settings: (i) terminal-state prediction and (ii) data-level scaling. For terminal-state prediction, we train for up to 200 epochs using the full training split. For data-level scaling, we train for up to 100 epochs on subsampled training sets to quantify sample efficiency. As a preprocessing step, we apply RevIN normalization to all inputs and targets, and report held-out test-set results in the normalized scale \citep{kim2021reversible}. 

\subsection{Terminal-state prediction}
Table~\ref{tab:compare_mse} reports terminal-state prediction accuracy (MSE) on the held-out test set for the two pretrained models on the terminal-state prediction task. On PLI, MORPH-Ti significantly outperforms POSEIDON-T ($0.055$ vs.\ $0.139$ MSE). On FRAC, POSEIDON-T achieves a modest advantage over MORPH-Ti ($0.037$ vs.\ $0.041$).

\begin{figure}[h]
    \centering
    \begin{subfigure}[b]{0.45\linewidth}
        \centering
        \includegraphics[trim={0cm 0cm 0cm 0cm},clip, width=1.0\textwidth]{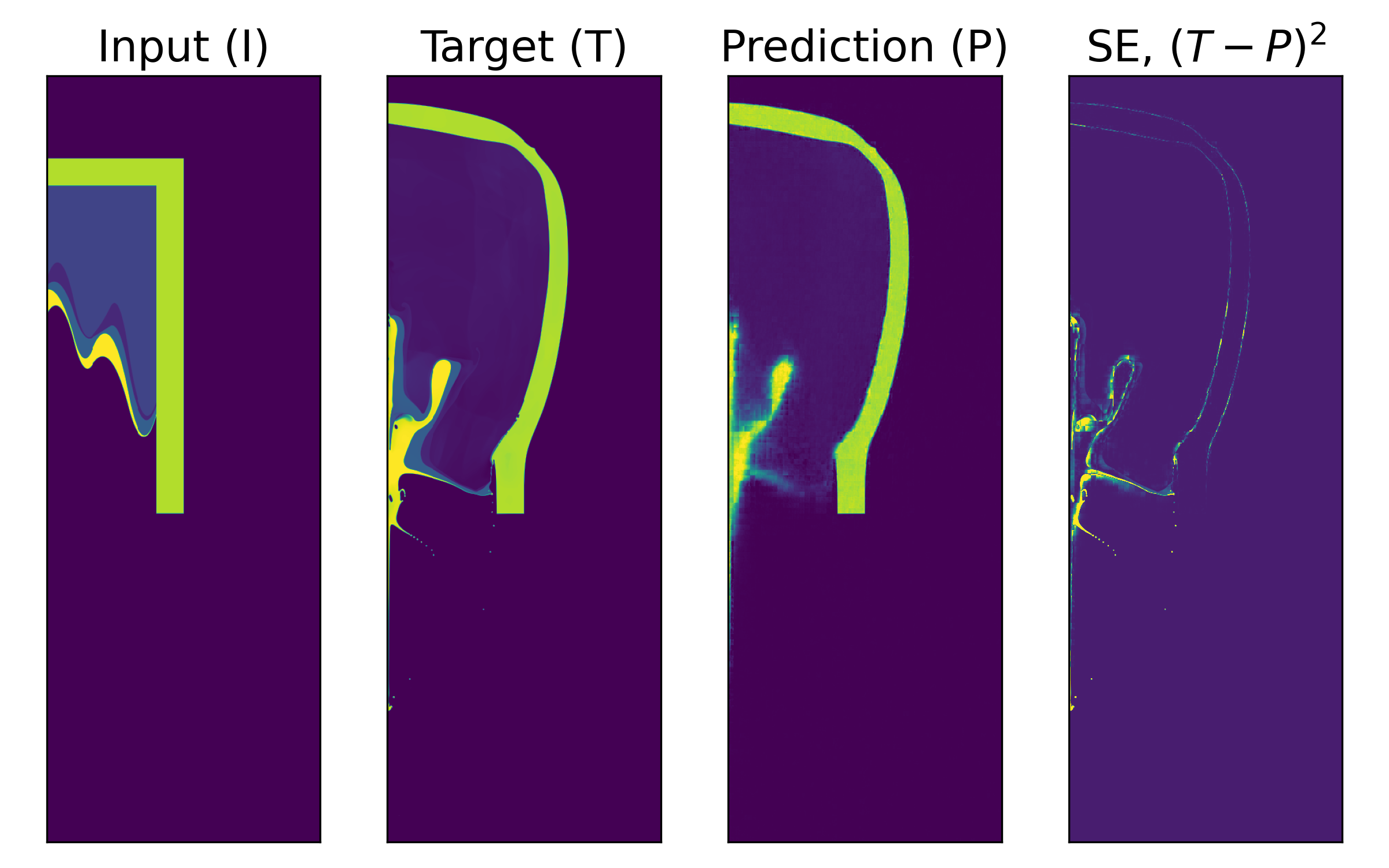}
    \end{subfigure}
    \hspace{1em}
    \begin{subfigure}[b]{0.45\linewidth}
        \centering
        \includegraphics[trim={0cm 0cm 0cm 0cm},clip, width=1.0\textwidth]{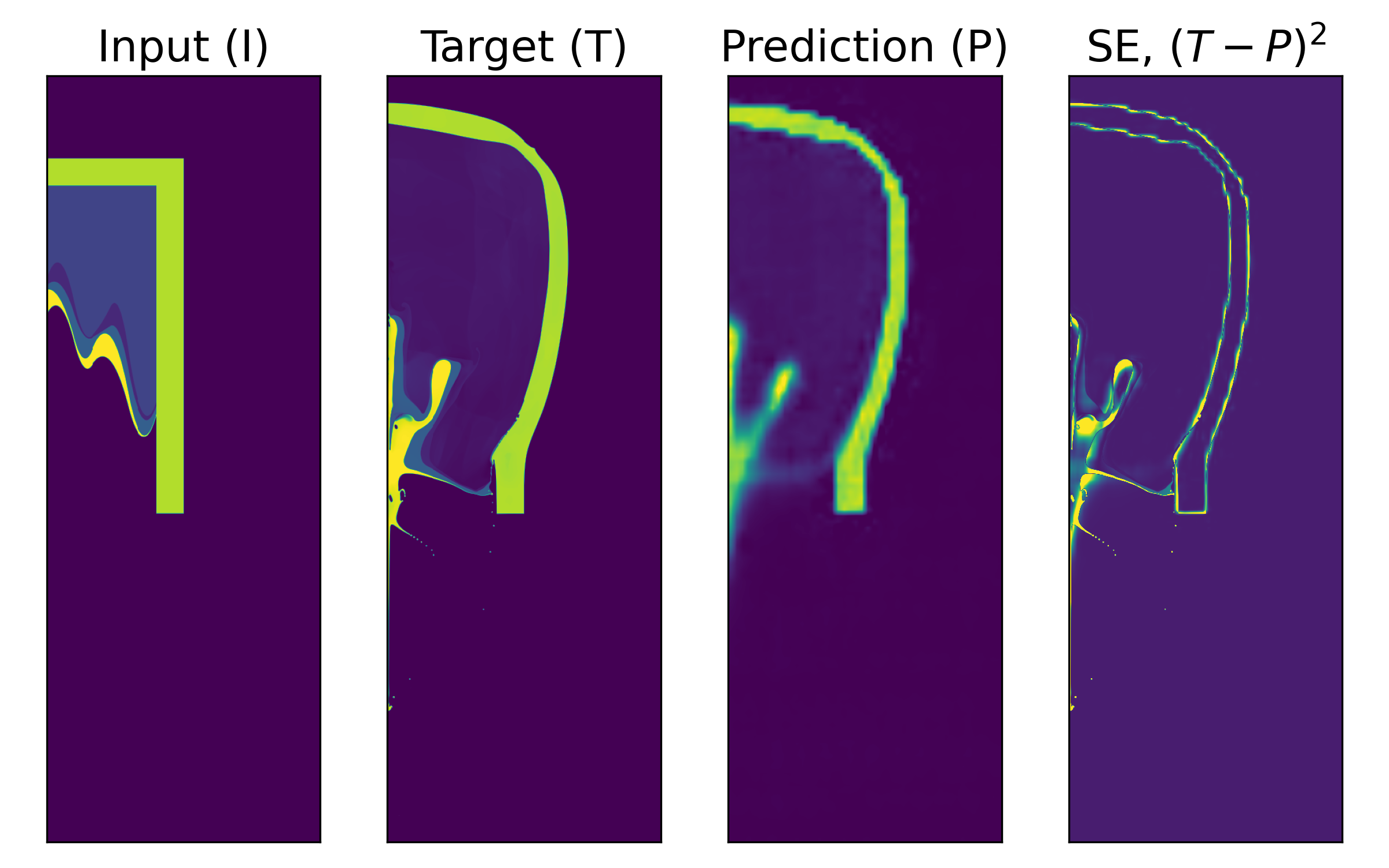}
    \end{subfigure}
    \begin{subfigure}[b]{0.45\linewidth}
        \centering
        \includegraphics[trim={0cm 0cm 0cm 0cm},clip, width=1.0\textwidth]{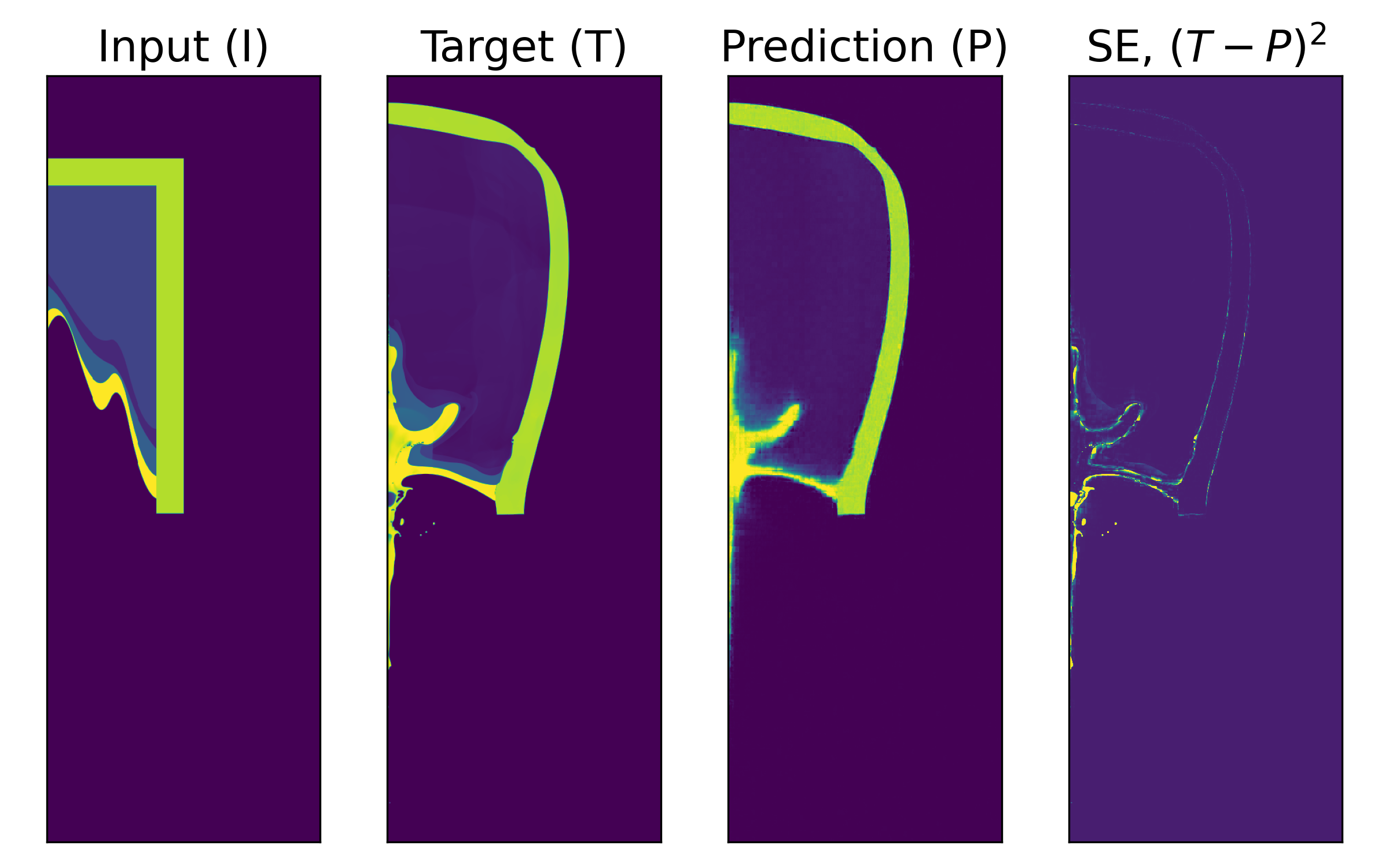}
        \caption{MORPH-Ti}
    \end{subfigure}
    \hspace{1em}
    \begin{subfigure}[b]{0.45\linewidth}
        \centering
        \includegraphics[trim={0cm 0cm 0cm 0cm},clip, width=1.0\textwidth]{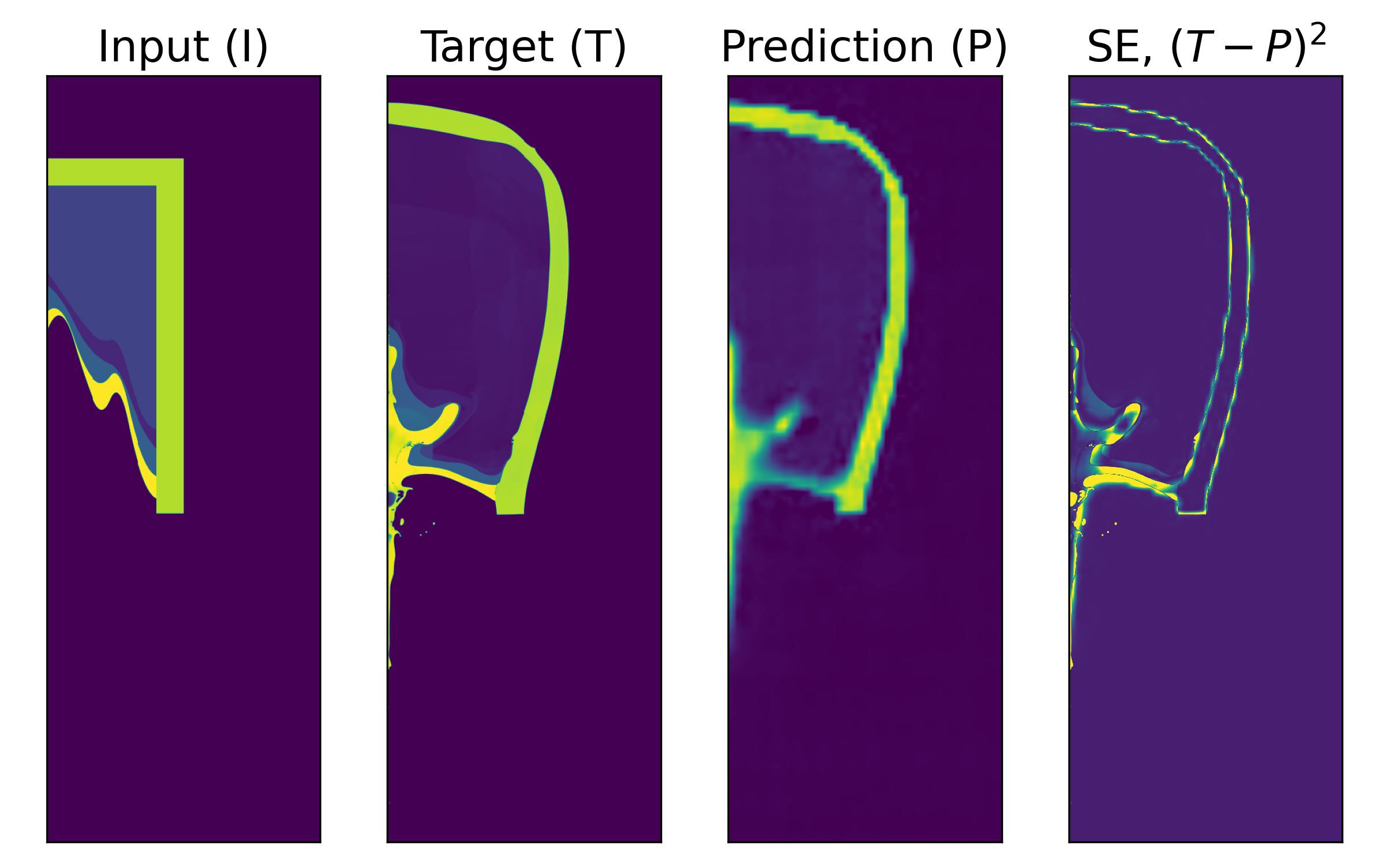}
        \caption{Poseidon-T}
    \end{subfigure}
    \caption{Terminal-state prediction on PLI comparing MORPH-Ti (a) and POSEIDON-T (b). Columns show the input, target terminal state, prediction, and squared error $(T-P)^2$.}
    \label{fig:heat_compare}
\end{figure}

\begin{figure}[h]
    \centering
    \begin{subfigure}[b]{0.45\linewidth}
        \centering
        \includegraphics[trim={0cm 0cm 0cm 0cm},clip, width=1.0\textwidth]{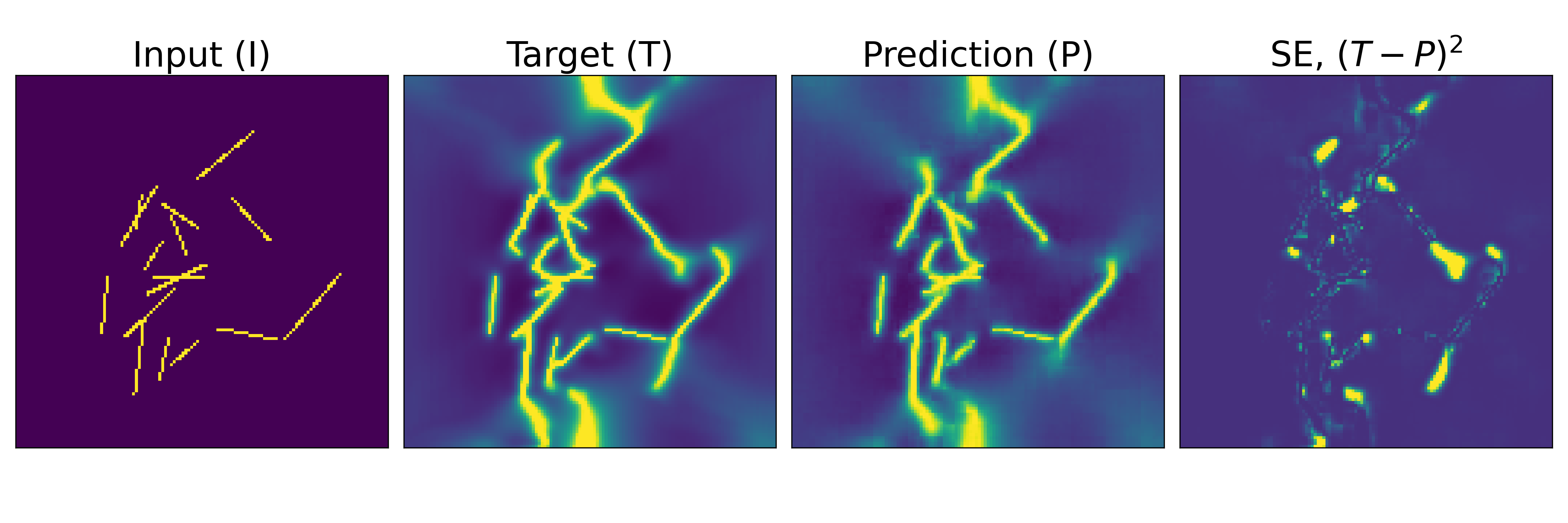}
    \end{subfigure}
    \hspace{1em}
    \begin{subfigure}[b]{0.45\linewidth}
        \centering
        \includegraphics[trim={0cm 0cm 0cm 0cm},clip, width=1.0\textwidth]{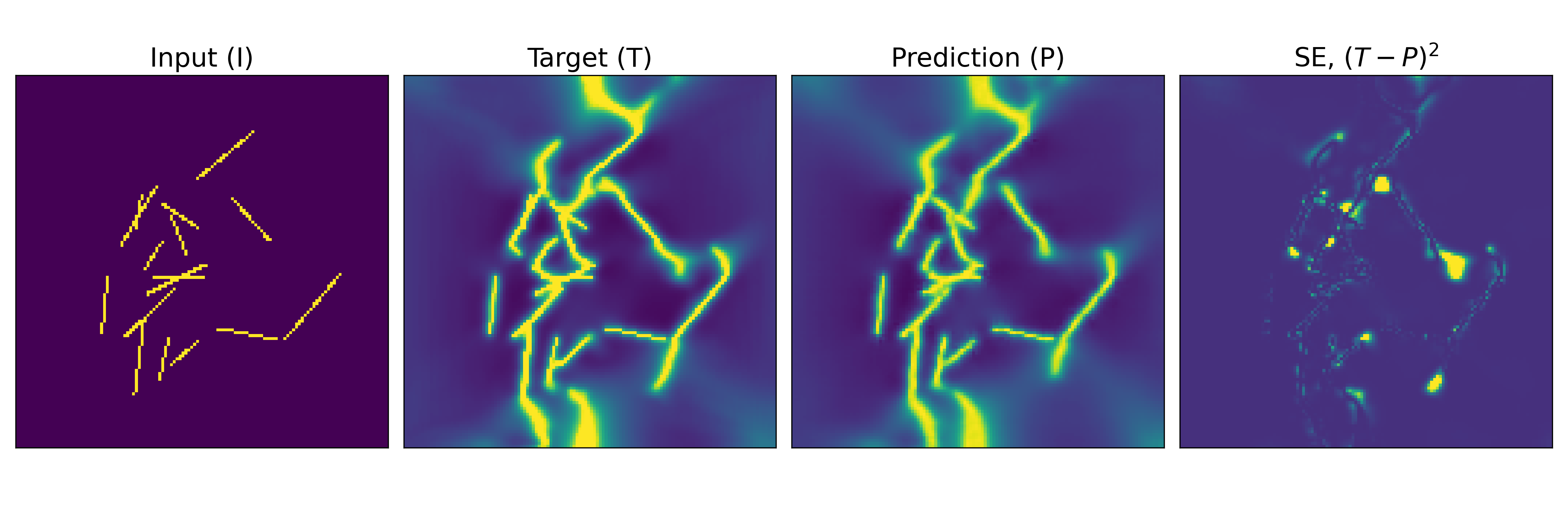}
    \end{subfigure}
    \begin{subfigure}[b]{0.45\linewidth}
        \centering
        \includegraphics[trim={0cm 0cm 0cm 0cm},clip, width=1.0\textwidth]{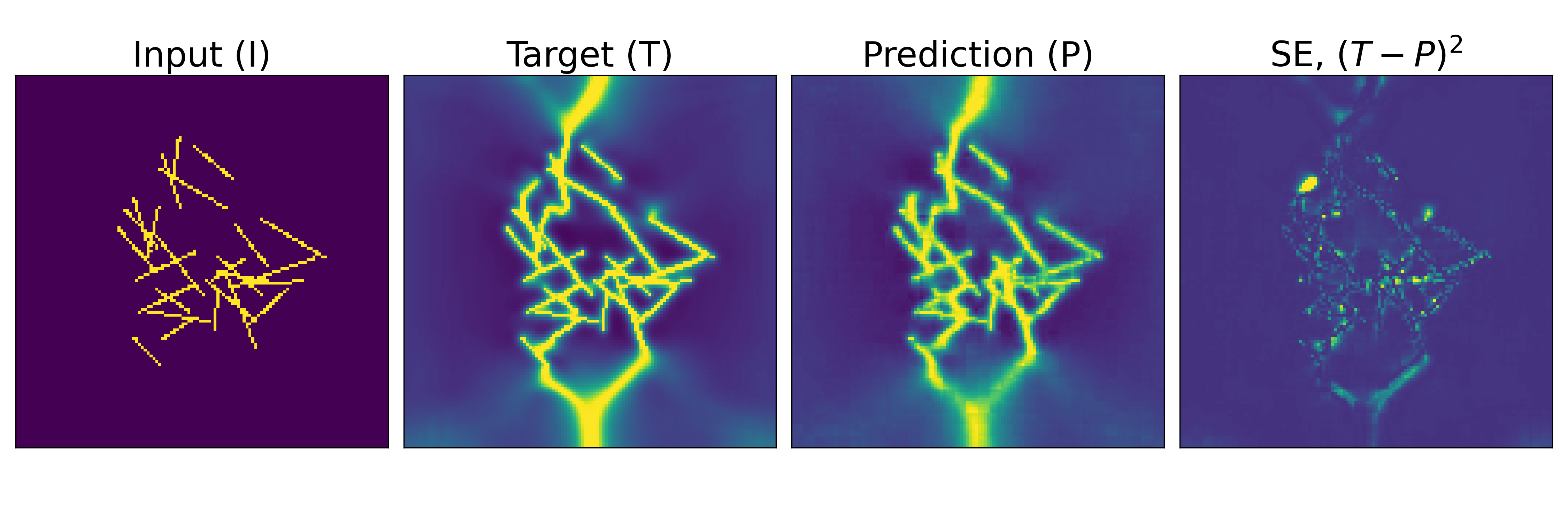}
        \caption{MORPH-Ti}
    \end{subfigure}
    \hspace{1em}
    \begin{subfigure}[b]{0.45\linewidth}
        \centering
        \includegraphics[trim={0cm 0cm 0cm 0cm},clip, width=1.0\textwidth]{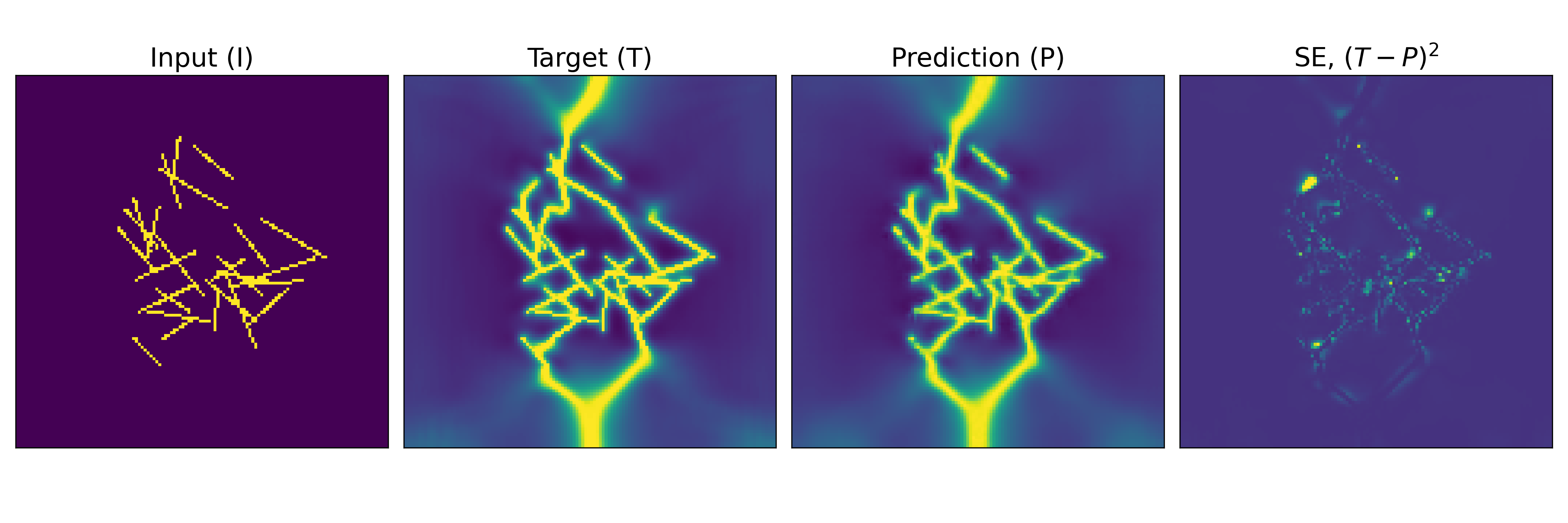}
        \caption{Poseidon-T}
    \end{subfigure}
    \caption{Terminal-state prediction on FRAC comparing MORPH-Ti (a) and POSEIDON-T (b). For each test example, columns show the input (initial state), target terminal state, model prediction, and squared error map $(T-P)^2$.}
    \label{fig:frac_compare}
\end{figure}

\begin{wraptable}{r}{0.5\textwidth} 
\vspace{-5pt}
\centering
\begin{tabular}{|c|c|c|}
\hline
\textbf{Dataset/Models} & \textbf{MORPH} & \textbf{POSEIDON} \\
\hline
PLI & 0.05445 & 0.13892 \\
\hline
FRAC & 0.04103 & 0.03704 \\
\hline
\end{tabular}
\caption{MSE on the test set}
\vspace{-10pt}
\label{tab:compare_mse}
\end{wraptable}
Figs.~\ref{fig:heat_compare} and \ref{fig:frac_compare} provide qualitative comparisons on two PLI and FRAC test examples, visualizing the input (initial state), the target terminal state, the model prediction, and the squared error map $(T-P)^2$.

On PLI, (Fig.~\ref{fig:heat_compare}), both models recover the coarse features, including the dominant interface displacement induced by the shock. Consistent with Table~1, MORPH-Ti tracks the terminal interface geometry more faithfully relative to POSEIDON-T. Additional qualitative examples are provided in Figs.~\ref{fig:heat_compare_ex1} and \ref{fig:heat_compare_ex2} (Appendix~\ref{app:results}).

On FRAC (Fig.~\ref{fig:frac_compare}), both models recover the overall terminal damage field. Residual errors are concentrated near crack tips, along thin branches, and at junctions where branching or coalescence occurs. Both models achieve comparable fidelity, with a small quantitative advantage for POSEIDON-T as reflected in Table~\ref{tab:compare_mse}. Additional examples are shown in Fig.~\ref{fig:frac_compare_ex} (Appendix~\ref{app:results}).

\subsection{Data-level Scaling}
\begin{figure}[h]
    \centering
    \begin{subfigure}[b]{0.45\linewidth}
        \centering
        \includegraphics[trim={0cm 0cm 0cm 0cm},clip, width=1.0\textwidth]{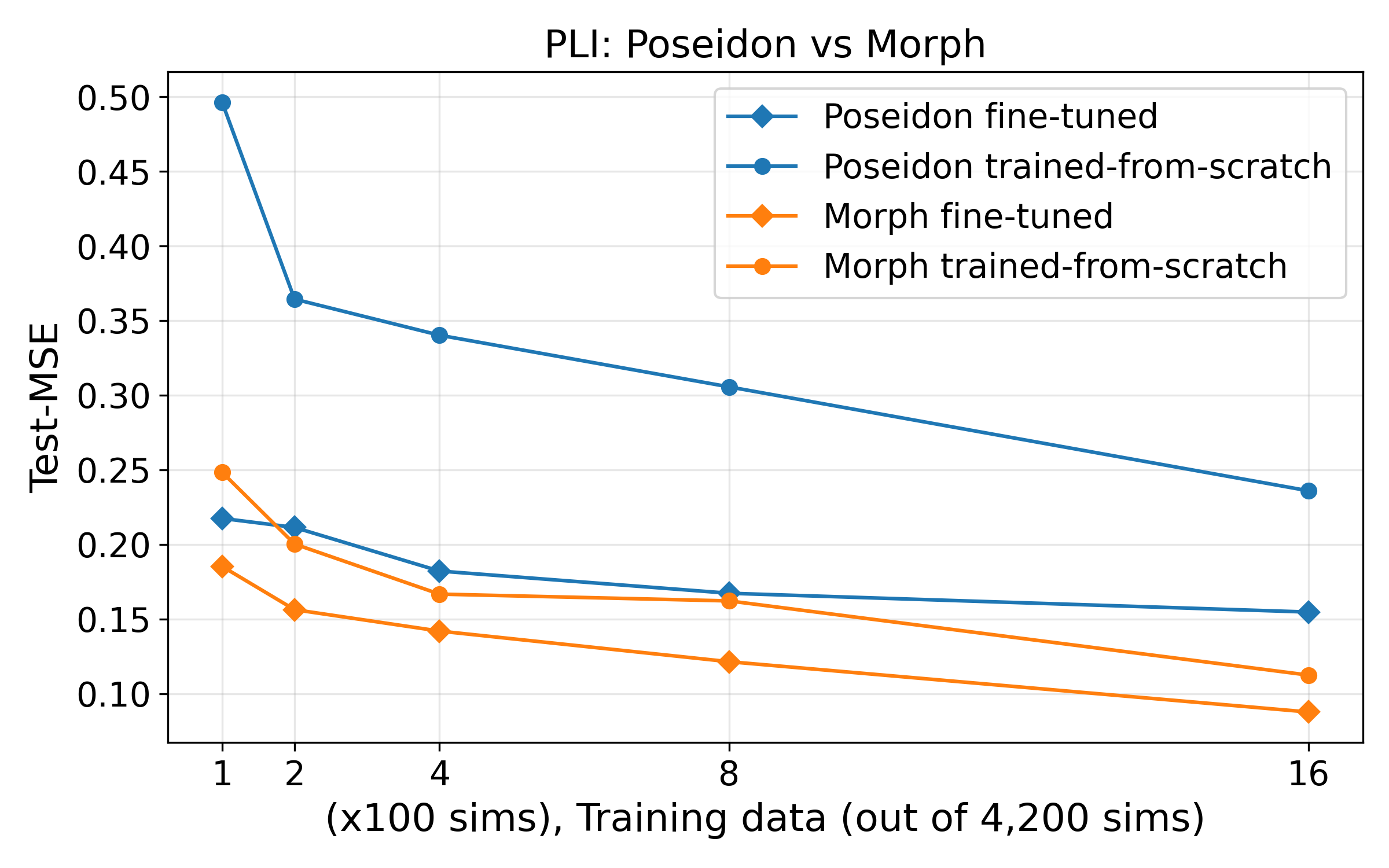}
        \caption{PLI}
    \end{subfigure}
    \hspace{1em}
    \begin{subfigure}[b]{0.45\linewidth}
        \centering
        \includegraphics[trim={0cm 0cm 0cm 0cm},clip, width=1.0\textwidth]{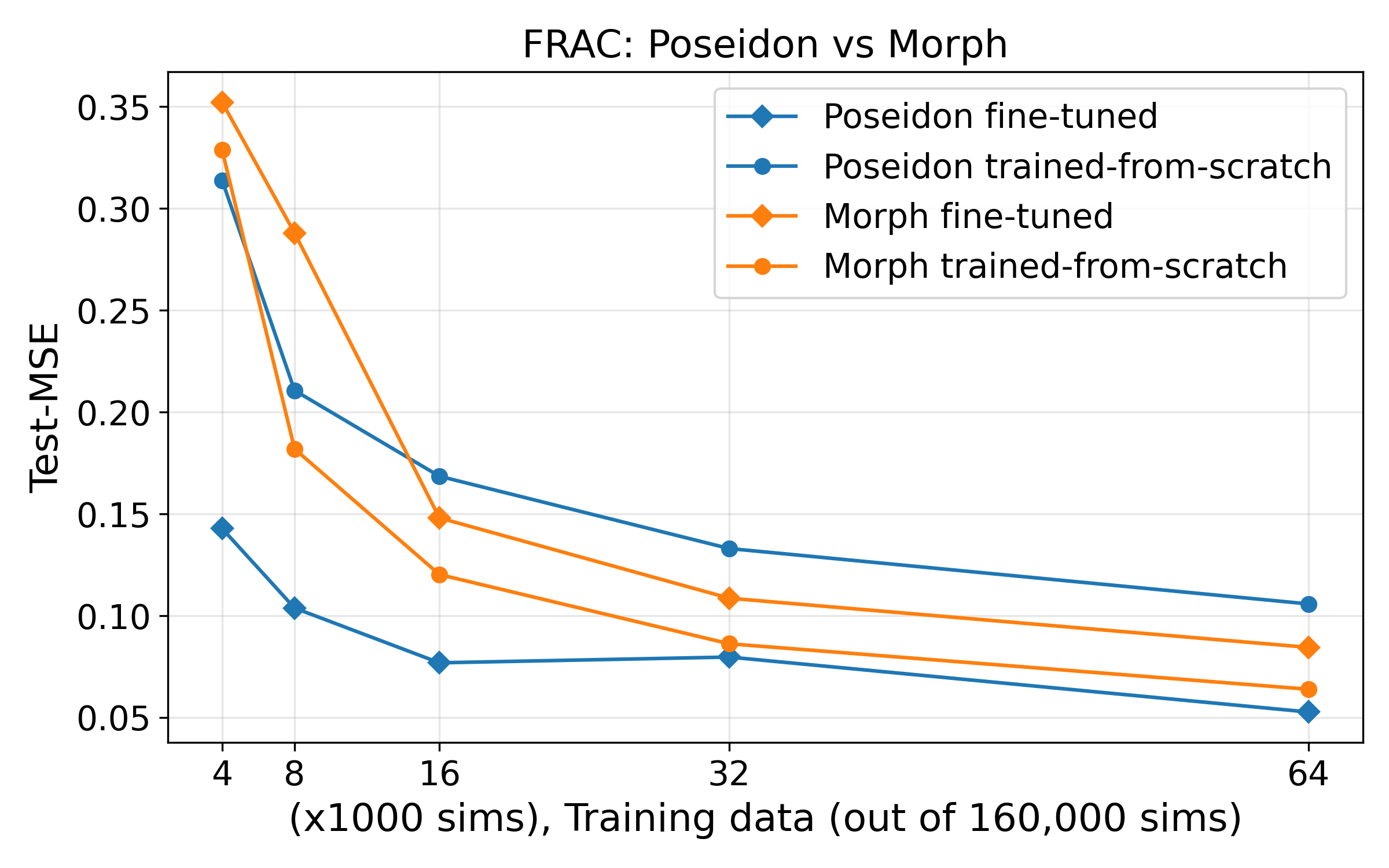}
        \caption{FRAC}
    \end{subfigure}
    \caption{Test MSE as a function of training-set size for PLI (a) and FRAC (b), comparing fine-tuning from pretrained weights versus training from scratch for MORPH and POSEIDON. Curves quantify sample efficiency for first-frame $\rightarrow$ final-frame operator learning.}
    \label{fig:scaling}
\end{figure}

To quantify sample efficiency, we perform data-level scaling studies on both PLI and FRAC. For each dataset, we vary the number of training simulations and compare (i) fine-tuning from pretrained weights and (ii) training the same architecture from random initialization. Fig.~\ref{fig:scaling} reports test error as a function of training-set size for PLI (subsamples of the full 4,200 simulations) and FRAC (subsamples of the full 160,000 simulations).

Across both regimes, increasing training data yields monotonically lower terminal-state error, but the rate of improvement depends strongly on initialization. Pretraining is most beneficial in the low-data regime, where fine-tuned models achieve lower error at matched data budgets, indicating improved sample efficiency under distribution shift. As the training set increases, the performance gap between fine-tuning and training from scratch shrinks, indicating that more in-domain data reduces the benefit of pretraining.

For PLI (Fig.~\ref{fig:scaling}a), we sweep the training set size from $1 \times 10^2$ to $16 \times 10^2$ simulations. MORPH attains substantially lower error than POSEIDON throughout the range. This is consistent with Table~\ref{tab:compare_mse} suggesting a more favorable inductive bias for shock-driven, interface-dominated dynamics. Fine-tuning also provides a clear advantage for POSEIDON in the low-data regime relative to training from scratch, whereas MORPH shows a smaller fine-tuning gap and the two MORPH curves converge as the dataset size increases. Notably, MORPH trained from scratch outperforms fine-tuned POSEIDON across the sweep.

For FRAC (Fig.~\ref{fig:scaling}b), we sweep from $4 \times 10^3$ to $64 \times 10^3$ simulations. Test error decreases with data for all four model configurations. POSEIDON benefits consistently from pretraining across budgets, with the largest improvement at smaller training sets and diminishing gains as data increases, consistent with improved sample efficiency from pretraining. In contrast, MORPH exhibits the opposite trend in this regime, with training from scratch outperforming fine-tuning at matched budgets, indicating that POSEIDON’s pretraining transfers more effectively to FRAC over the explored data range. Despite this negative transfer, MORPH trained from scratch matches the performance of fine-tuned POSEIDON on FRAC.

Compared to the transfer gains reported across 10 fine-tuning benchmarks in \citep{rautela2025morph} and 15 benchmarks in \citep{herde2024poseidon}, the advantage of fine-tuning over training from scratch is smaller in our extreme-loading setting. Even in the low-data regime, the transfer gain (defined at a fixed training-set size as the ratio of test loss for the trained-from-scratch model to that of the fine-tuned model) peaks at only $\approx 2\times$. This modest gain motivates incorporating extreme-loading material dynamics into future pretraining corpora to boost transfer gains on shock- and fracture-dominated datasets. Continual learning may provide an additional pathway for integrating such new datasets without retraining models from scratch. Such an approach could enable models to progressively adapt to new physical regimes while preserving performance on previously learned material behaviors \citep{rolnick2019experience}.

\section{Conclusions}
We studied out-of-distribution transfer of PDE foundation models to extreme-loading material dynamics using two complementary, discontinuity-dominated regimes: shock-driven multi-material interface evolution (PLI) and dynamic fracture/failure evolution (FRAC). We formulated adaptation as terminal-state prediction (first-frame $\rightarrow$ final-frame) under a unified fine-tuning protocol, and evaluated two pretrained PDE-FMs: MORPH and POSEIDON against training from scratch across data budgets. Across datasets, transfer behavior is regime dependent: MORPH is more accurate on PLI, whereas POSEIDON attains a modest advantage on FRAC. Data-level scaling indicates that pretraining can improve sample efficiency in the low-data regime, but the benefit shrinks with additional in-domain supervision and can vary with the target physics and fine-tuning setup. Overall, these results suggest that fluid-centric pretraining alone may be insufficient to enable substantial transfer gains in shock- and fracture-dominated mechanics, underscoring the need for future PDE foundation models to incorporate more diverse extreme-loading datasets during pretraining. Continual learning strategies may further enable the progressive integration of such regimes without requiring full retraining of foundation models.

\section*{Acknowledgments}
Research presented in this article was supported by the Laboratory Directed Research and Development program of Los Alamos National Laboratory under project number 20250637DI. This research used resources provided by the Los Alamos National Laboratory Institutional Computing Program, which is supported by the U.S. Department of Energy National Nuclear Security Administration under Contract No. 89233218CNA000001.

\bibliography{iclr2026_conference}
\bibliographystyle{iclr2026_conference}

\appendix
\newpage
\section{Datasets} \label{app:dataset}

\begin{figure}[h]
\centering
\includegraphics[width=1.0\textwidth,trim=0cm 0cm 0cm 0cm, clip]{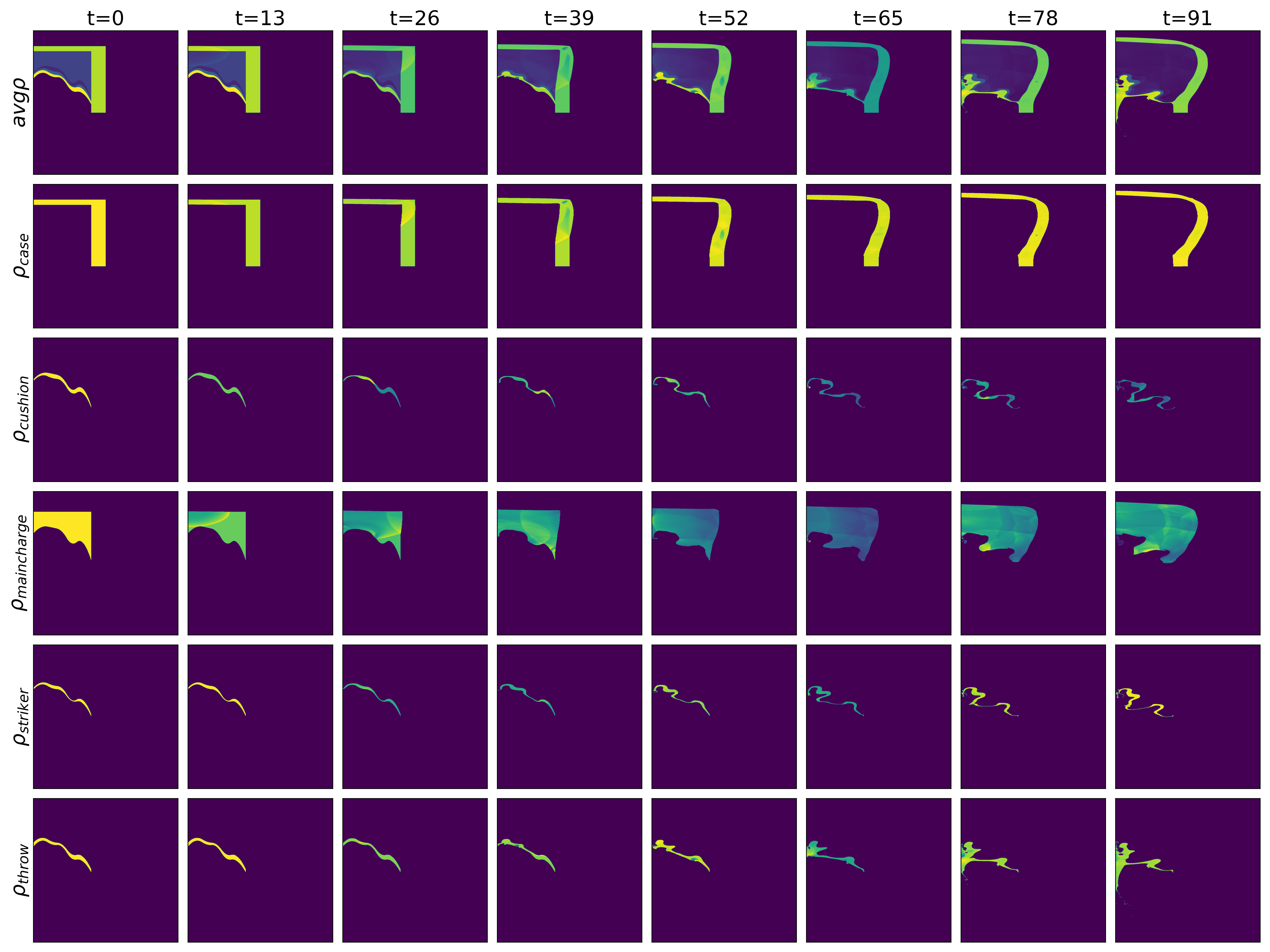}
\caption{Representative PLI spatiotemporal evolution over uniformly spaced time-steps. Rows visualize selected per-material density channels.}
\label{fig:pli}
\end{figure}

\begin{figure}[h]
    \centering
    \begin{subfigure}[b]{1.0\linewidth}
        \centering
        \includegraphics[trim={0cm 0cm 0cm 0cm},clip, width=1.0\textwidth]{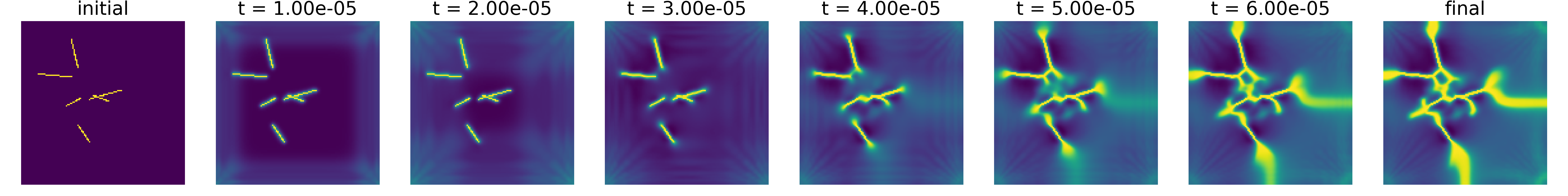}
    \end{subfigure}
    \begin{subfigure}[b]{1.0\linewidth}
        \centering
        \includegraphics[trim={0cm 0cm 0cm 0cm},clip, width=1.0\textwidth]{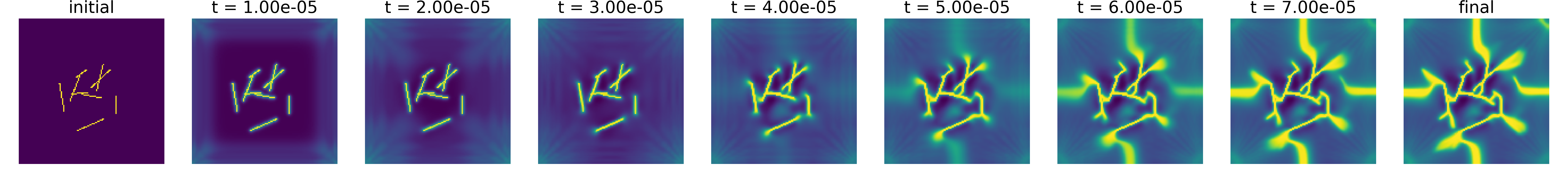}
    \end{subfigure}
    \begin{subfigure}[b]{1.0\linewidth}
        \centering
        \includegraphics[trim={0cm 0cm 0cm 0cm},clip, width=1.0\textwidth]{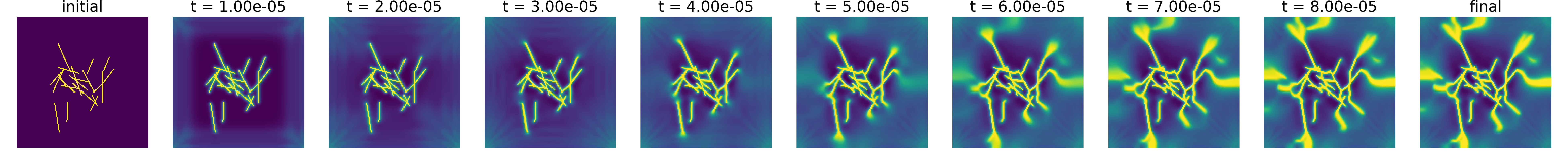}
    \end{subfigure}
    \caption{Representative FRAC trajectories (tungsten subset) showing the evolution of fracture/damage patterns from the initial condition to the terminal state. Each row corresponds to a distinct simulation, with frames annotated by characteristic time.}
    \label{fig:frac}
\end{figure}


\newpage
\section{Extended Results} \label{app:results}
\begin{figure}[h]
    \centering
    \begin{subfigure}[b]{0.45\linewidth}
        \centering
        \includegraphics[trim={0cm 0cm 0cm 0cm},clip, width=1.0\textwidth]{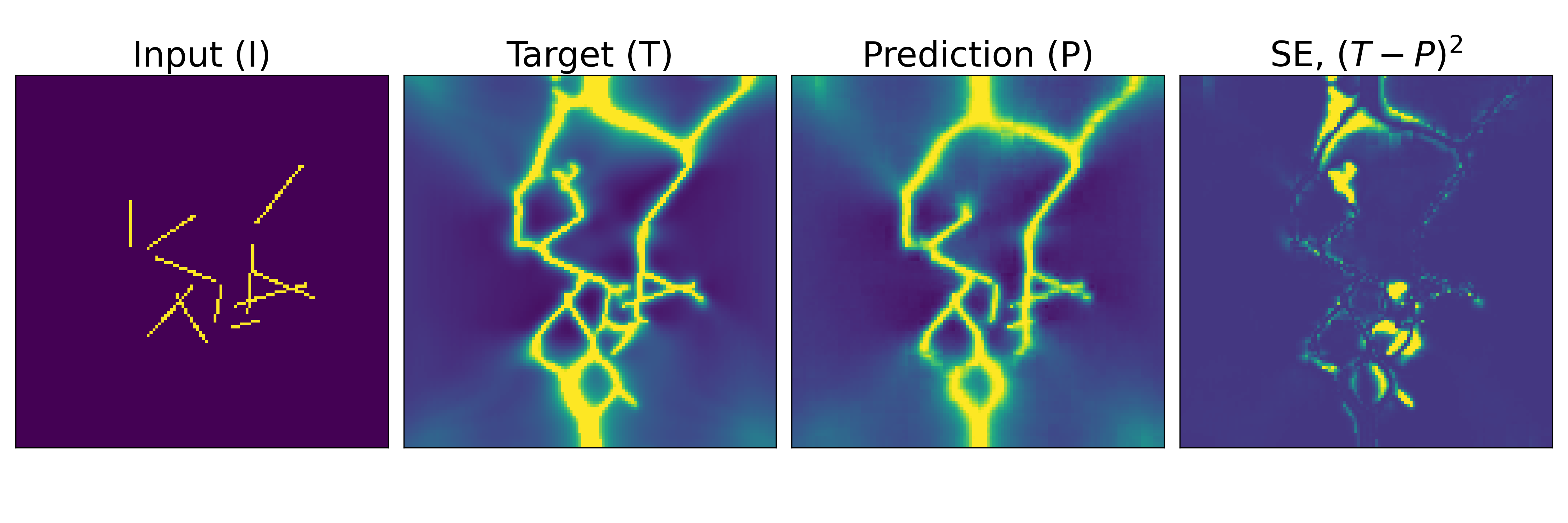}
    \end{subfigure}
    \hspace{1em}
    \begin{subfigure}[b]{0.45\linewidth}
        \centering
        \includegraphics[trim={0cm 0cm 0cm 0cm},clip, width=1.0\textwidth]{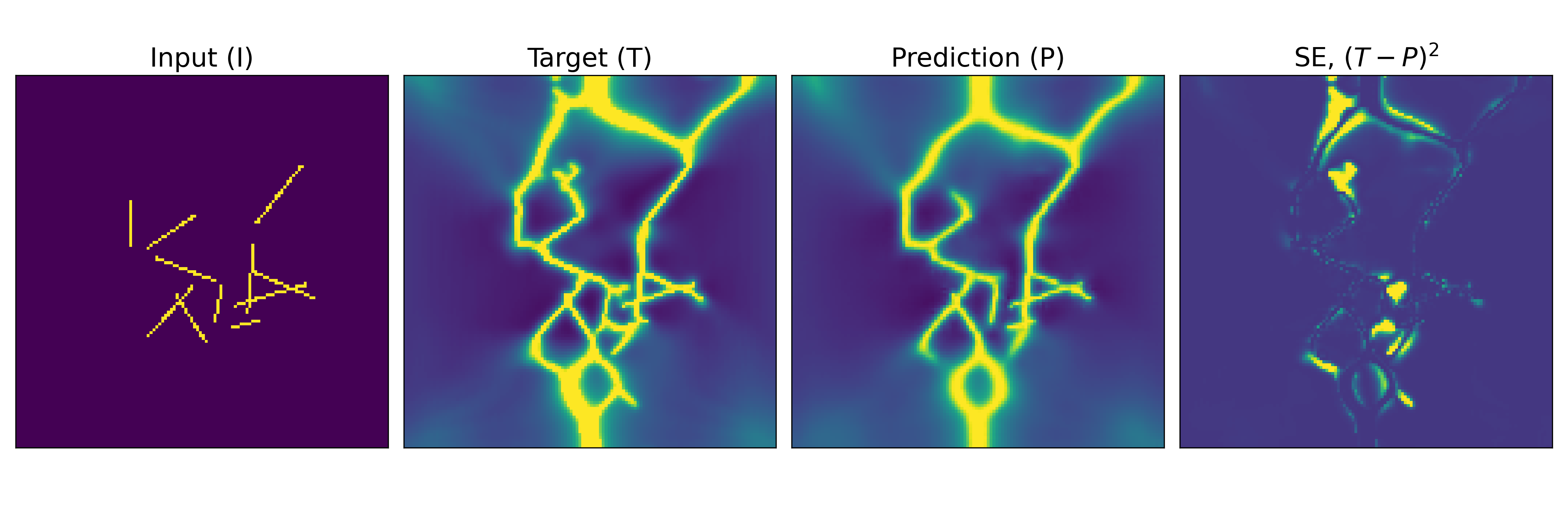}
    \end{subfigure}
    \begin{subfigure}[b]{0.45\linewidth}
        \centering
        \includegraphics[trim={0cm 0cm 0cm 0cm},clip, width=1.0\textwidth]{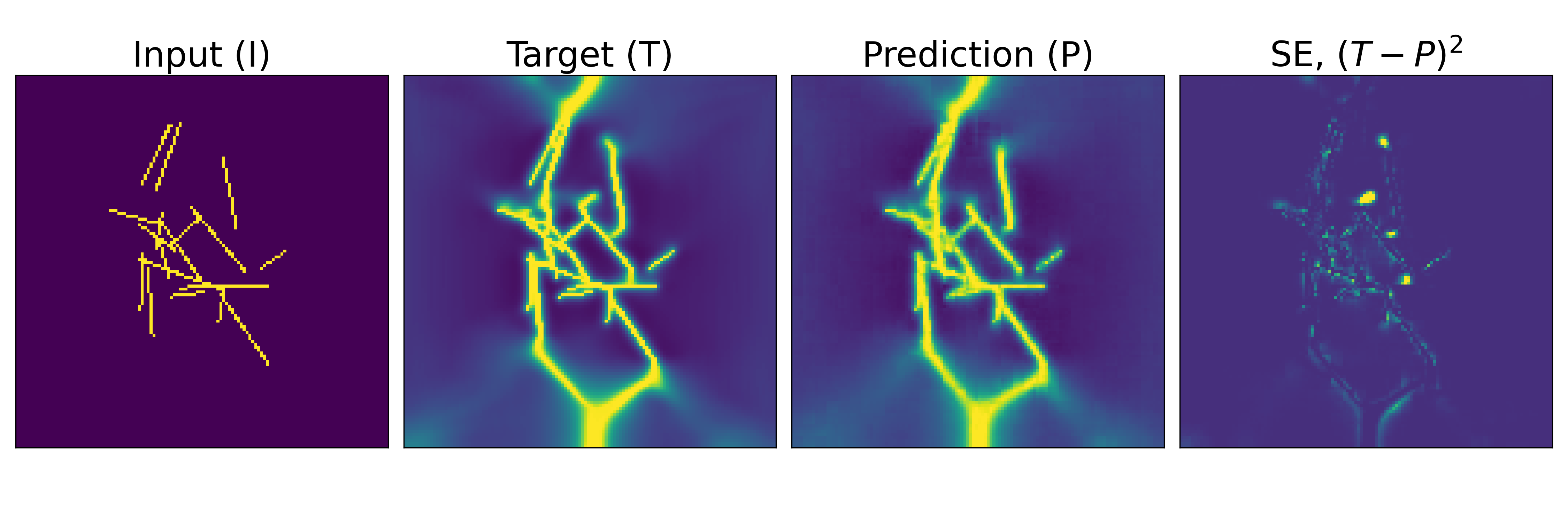}
    \end{subfigure}
    \hspace{1em}
    \begin{subfigure}[b]{0.45\linewidth}
        \centering
        \includegraphics[trim={0cm 0cm 0cm 0cm},clip, width=1.0\textwidth]{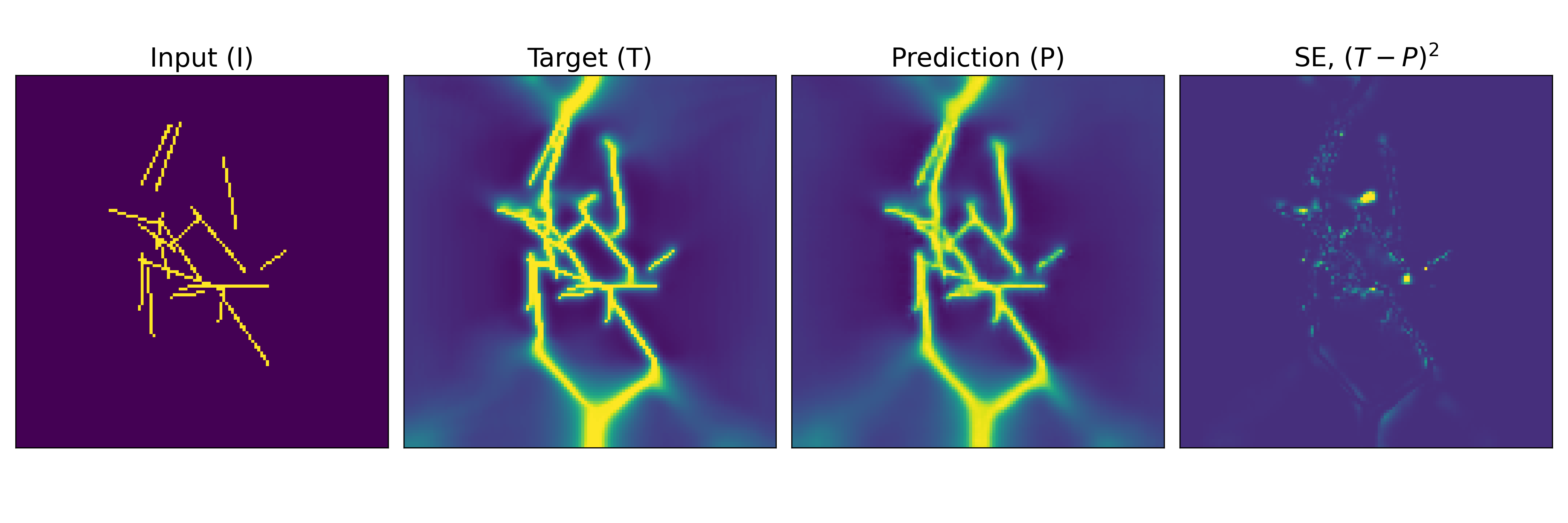}
    \end{subfigure}
    \begin{subfigure}[b]{0.45\linewidth}
        \centering
        \includegraphics[trim={0cm 0cm 0cm 0cm},clip, width=1.0\textwidth]{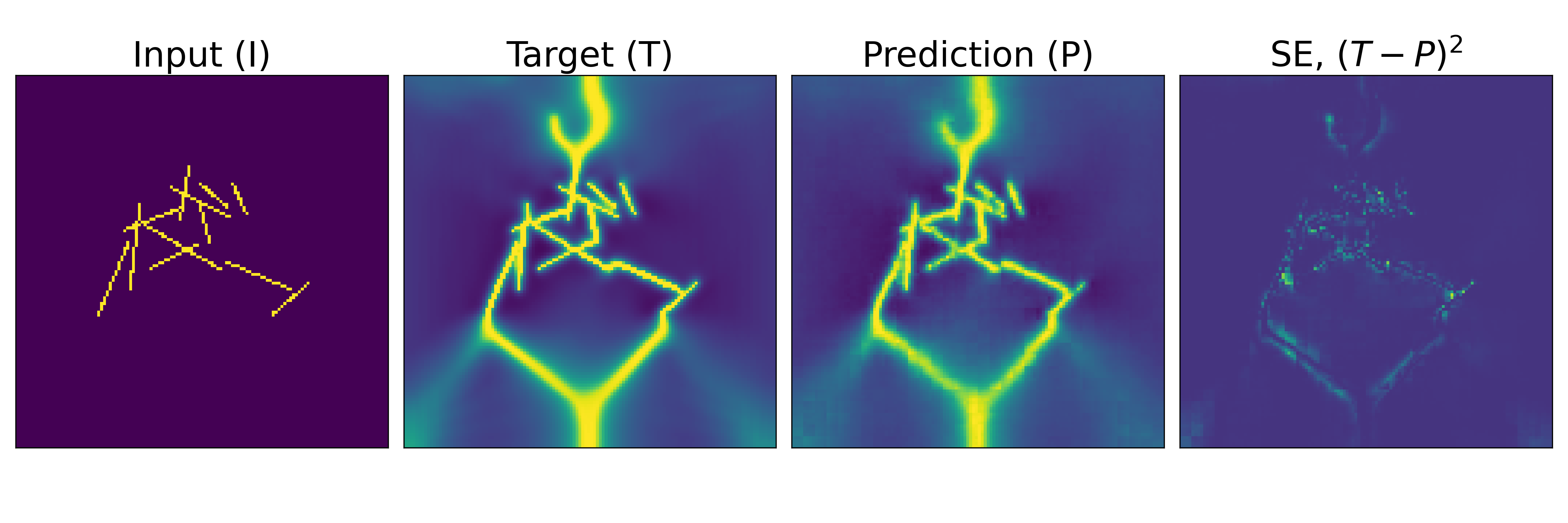}
    \end{subfigure}
    \hspace{1em}
    \begin{subfigure}[b]{0.45\linewidth}
        \centering
        \includegraphics[trim={0cm 0cm 0cm 0cm},clip, width=1.0\textwidth]{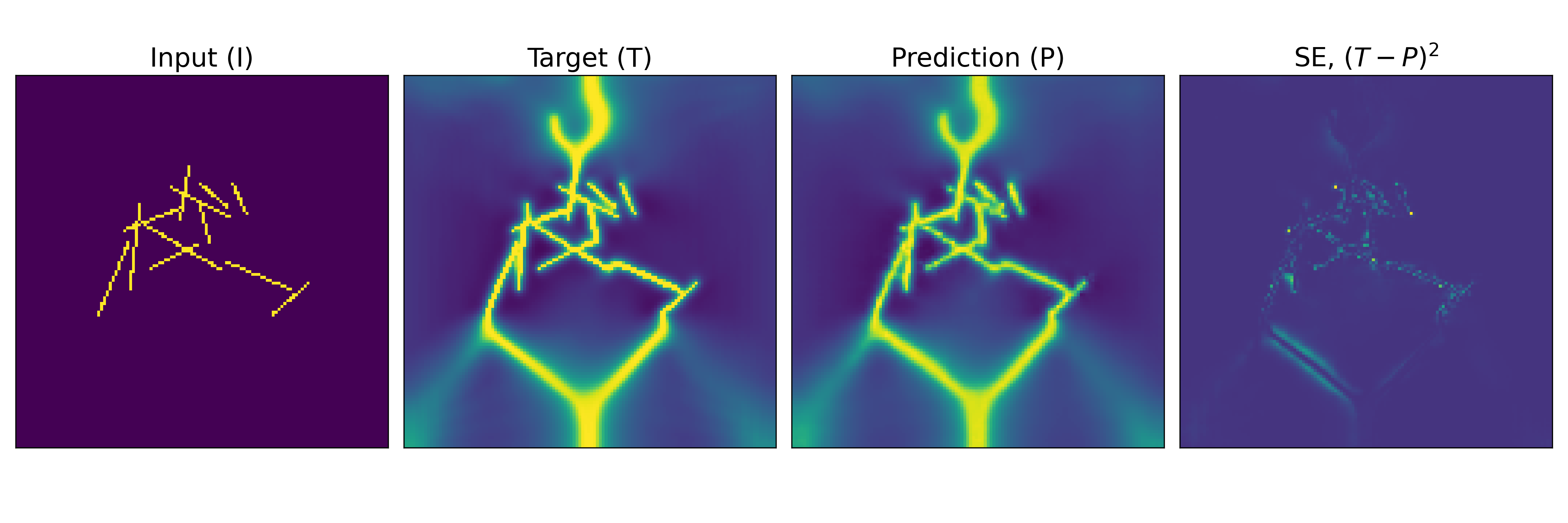}
    \end{subfigure}
    \begin{subfigure}[b]{0.45\linewidth}
        \centering
        \includegraphics[trim={0cm 0cm 0cm 0cm},clip, width=1.0\textwidth]{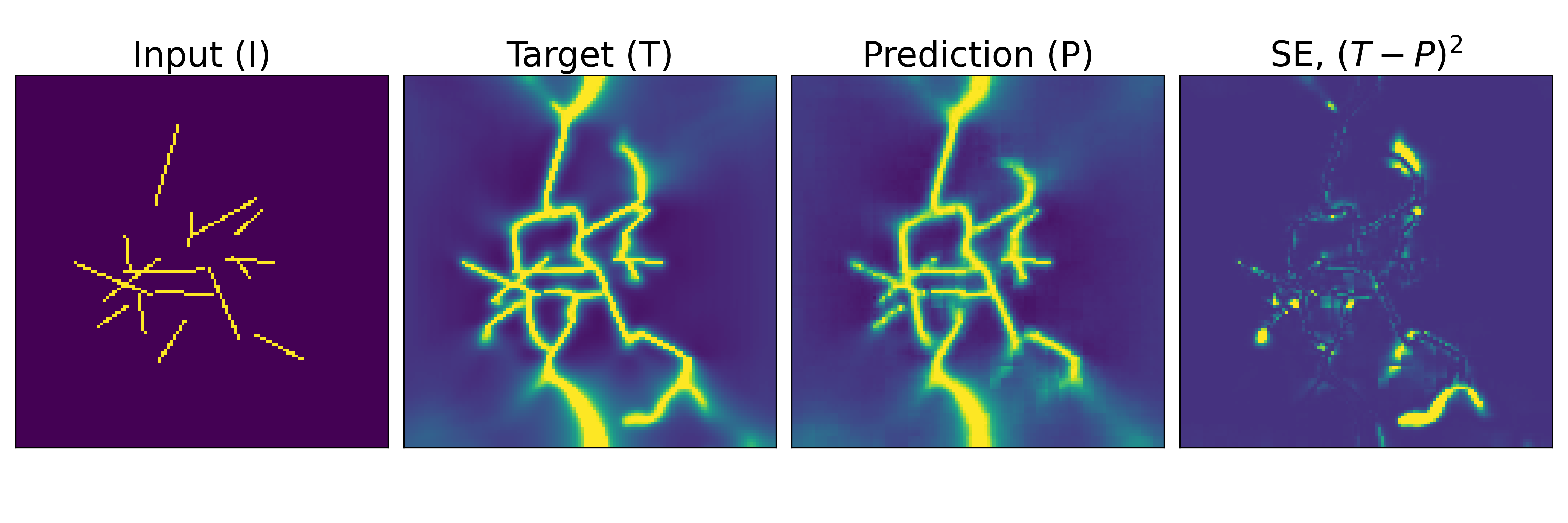}
    \end{subfigure}
    \hspace{1em}
    \begin{subfigure}[b]{0.45\linewidth}
        \centering
        \includegraphics[trim={0cm 0cm 0cm 0cm},clip, width=1.0\textwidth]{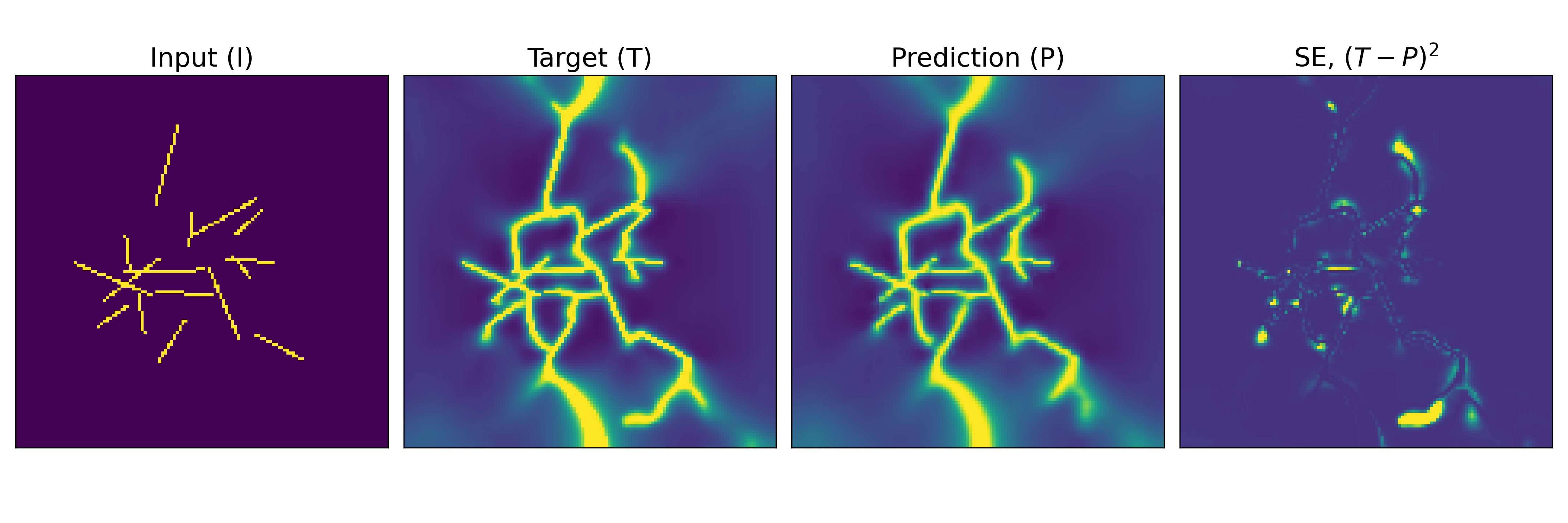}
    \end{subfigure}
    \begin{subfigure}[b]{0.45\linewidth}
        \centering
        \includegraphics[trim={0cm 0cm 0cm 0cm},clip, width=1.0\textwidth]{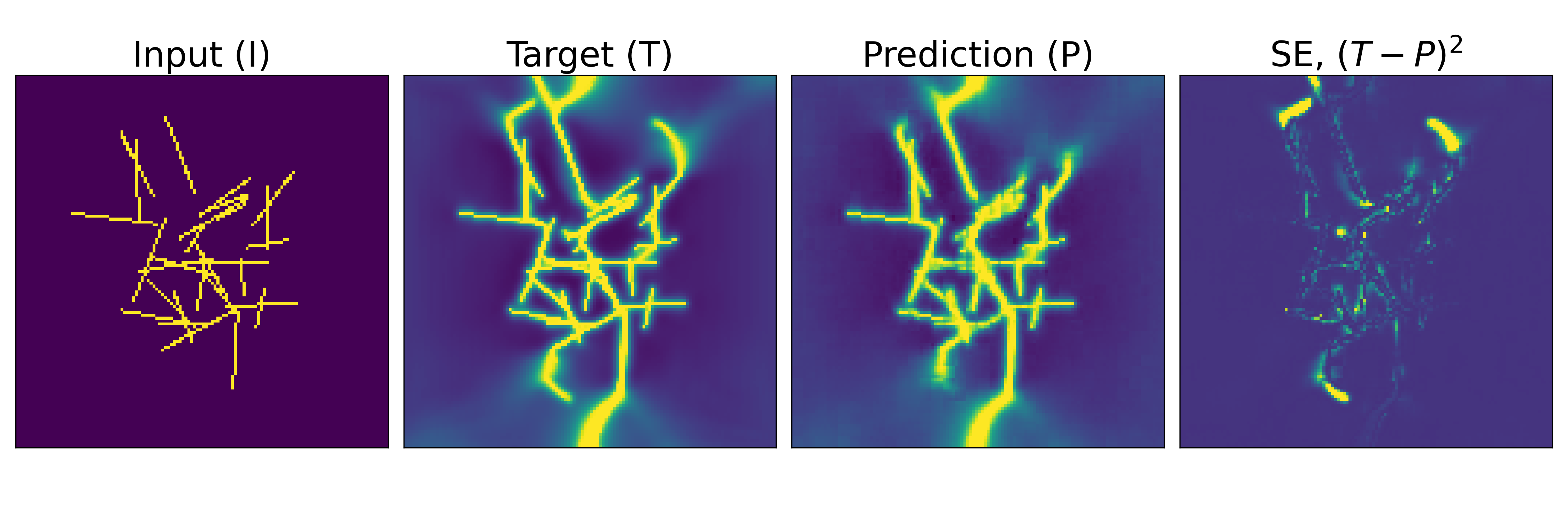}
    \end{subfigure}
    \hspace{1em}
    \begin{subfigure}[b]{0.45\linewidth}
        \centering
        \includegraphics[trim={0cm 0cm 0cm 0cm},clip, width=1.0\textwidth]{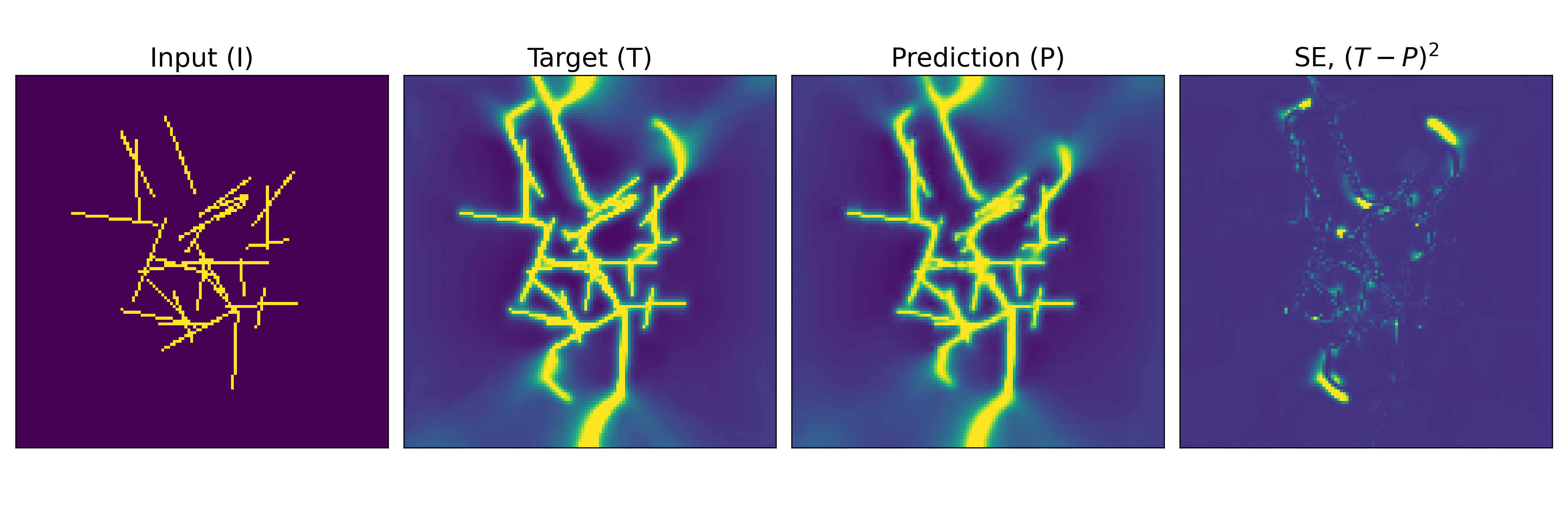}
    \end{subfigure}
    \begin{subfigure}[b]{0.45\linewidth}
        \centering
        \includegraphics[trim={0cm 0cm 0cm 0cm},clip, width=1.0\textwidth]{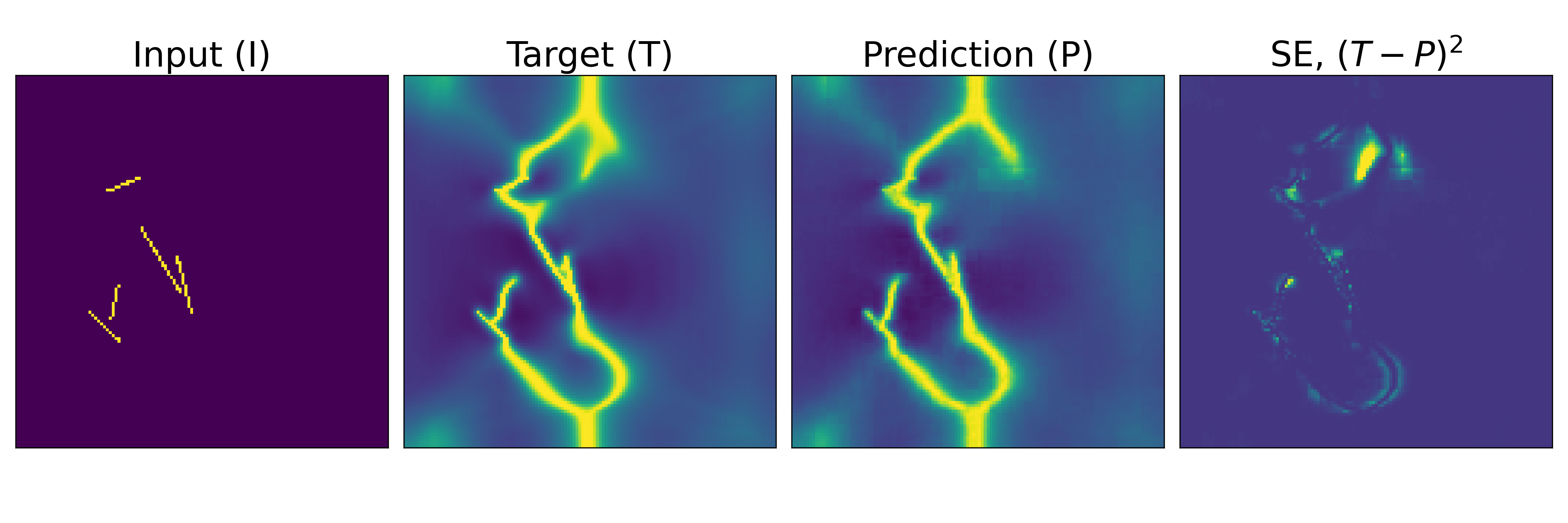}
    \end{subfigure}
    \hspace{1em}
    \begin{subfigure}[b]{0.45\linewidth}
        \centering
        \includegraphics[trim={0cm 0cm 0cm 0cm},clip, width=1.0\textwidth]{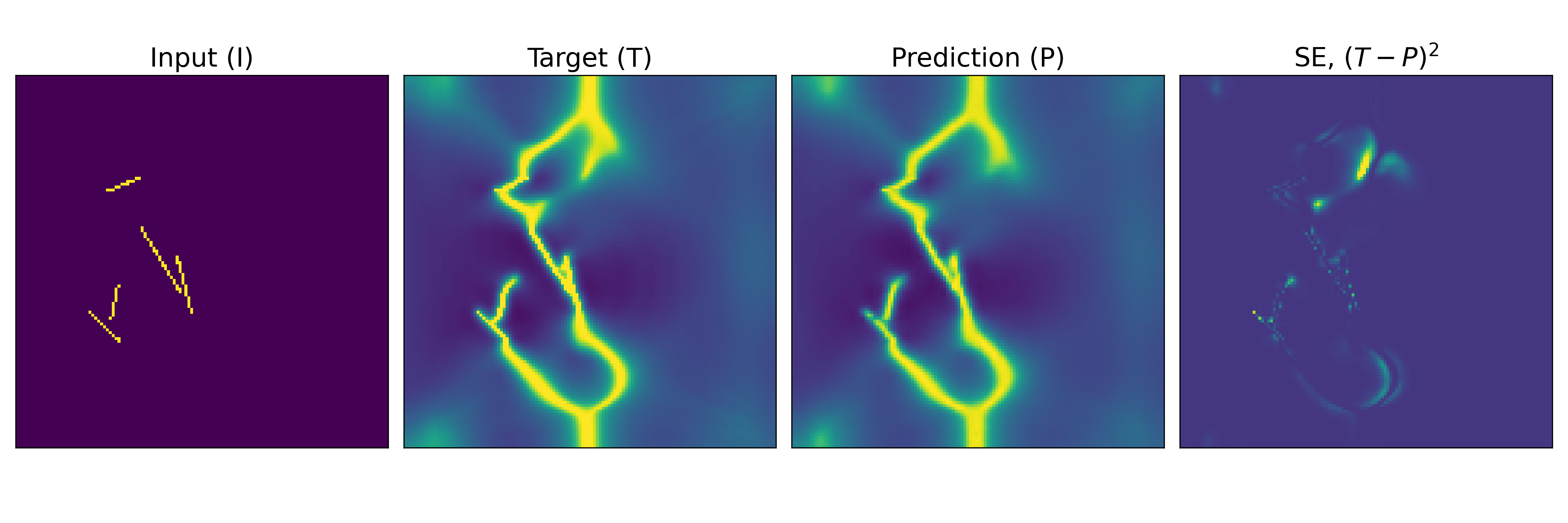}
    \end{subfigure}
    \begin{subfigure}[b]{0.45\linewidth}
        \centering
        \includegraphics[trim={0cm 0cm 0cm 0cm},clip, width=1.0\textwidth]{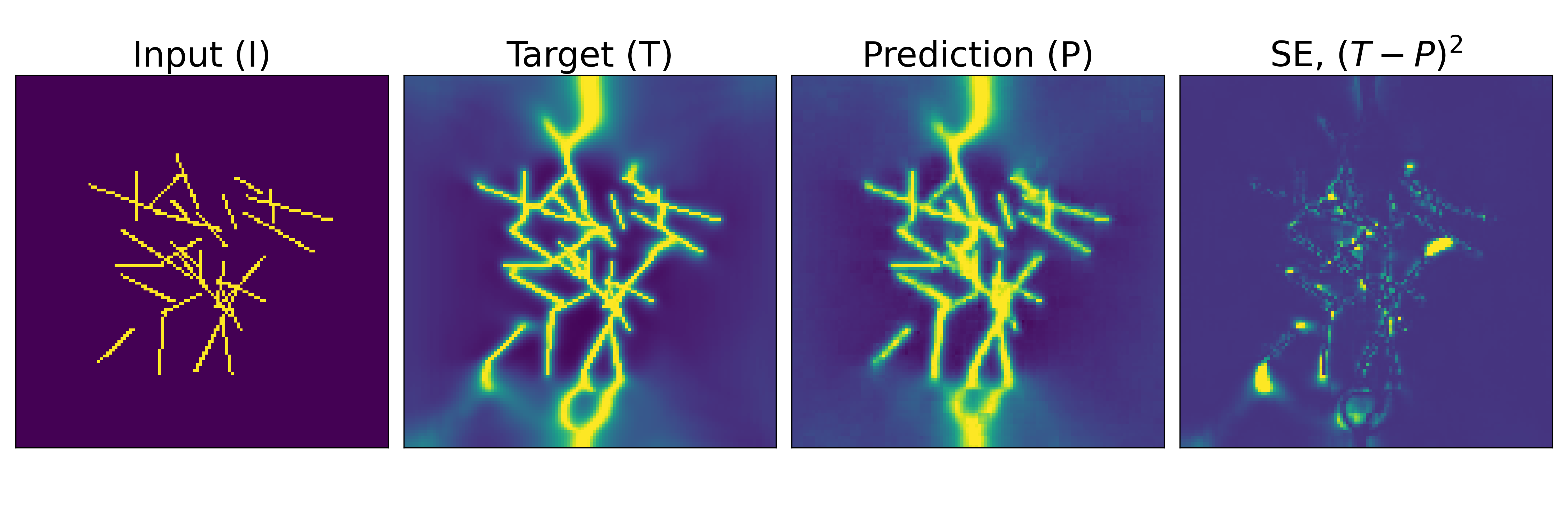}
    \end{subfigure}
    \hspace{1em}
    \begin{subfigure}[b]{0.45\linewidth}
        \centering
        \includegraphics[trim={0cm 0cm 0cm 0cm},clip, width=1.0\textwidth]{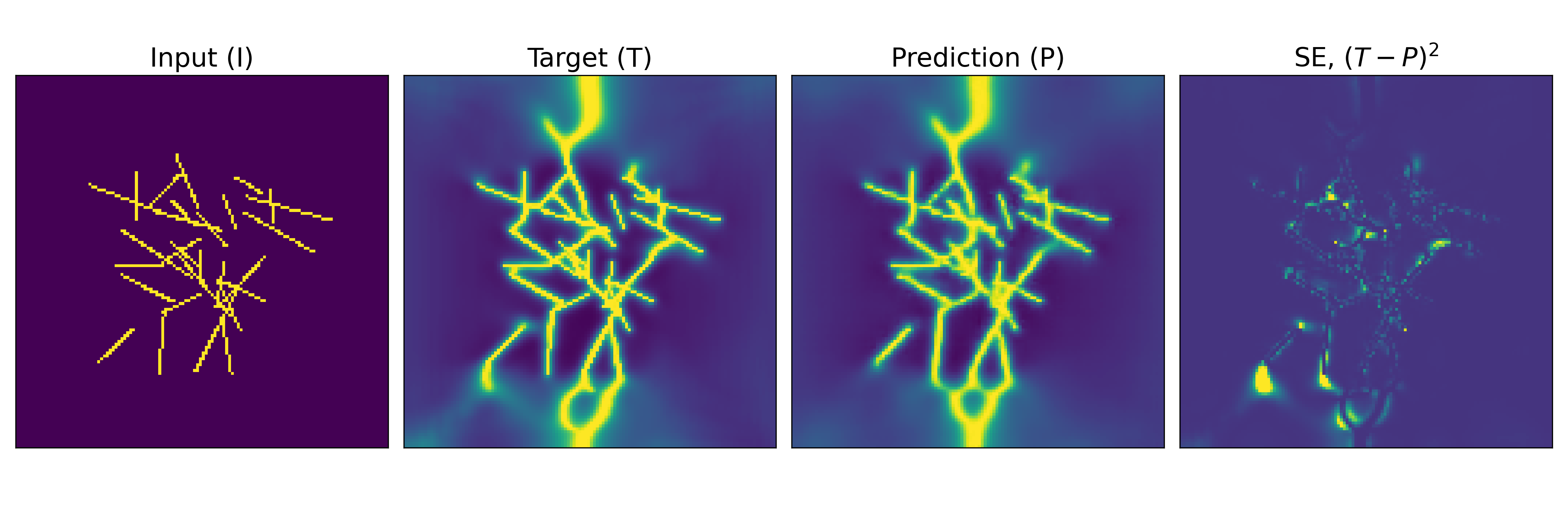}
    \end{subfigure}
    \begin{subfigure}[b]{0.45\linewidth}
        \centering
        \includegraphics[trim={0cm 0cm 0cm 0cm},clip, width=1.0\textwidth]{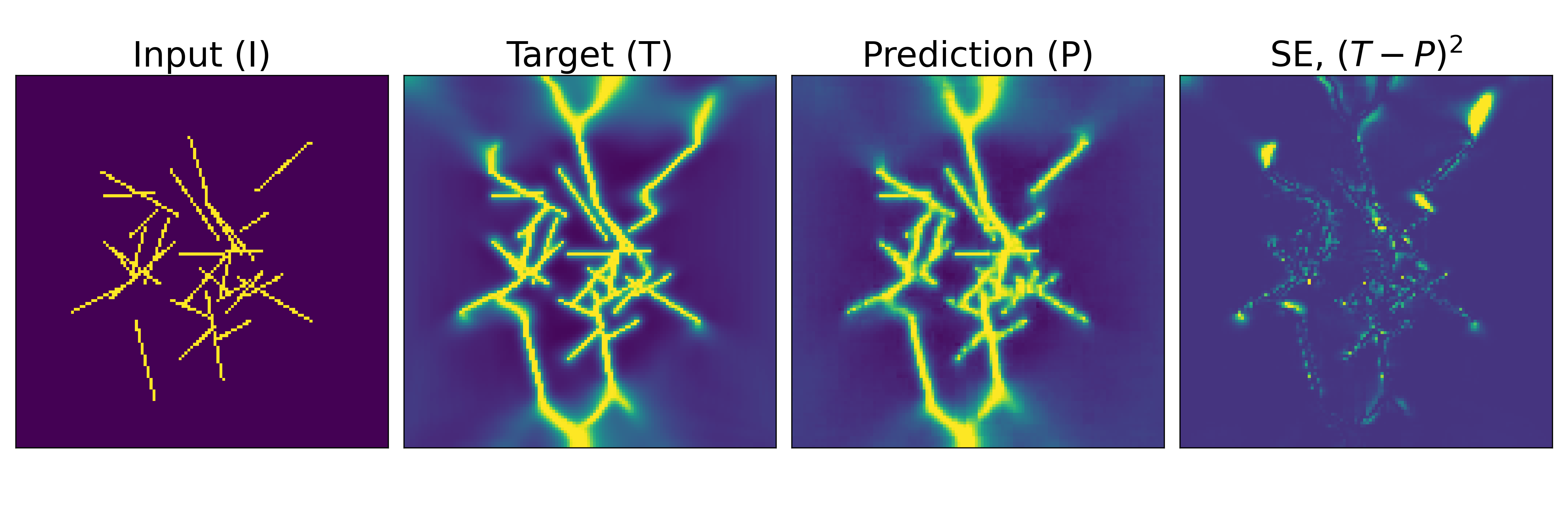}
        \caption{MORPH-Ti}
    \end{subfigure}
    \hspace{1em}
    \begin{subfigure}[b]{0.45\linewidth}
        \centering
        \includegraphics[trim={0cm 0cm 0cm 0cm},clip, width=1.0\textwidth]{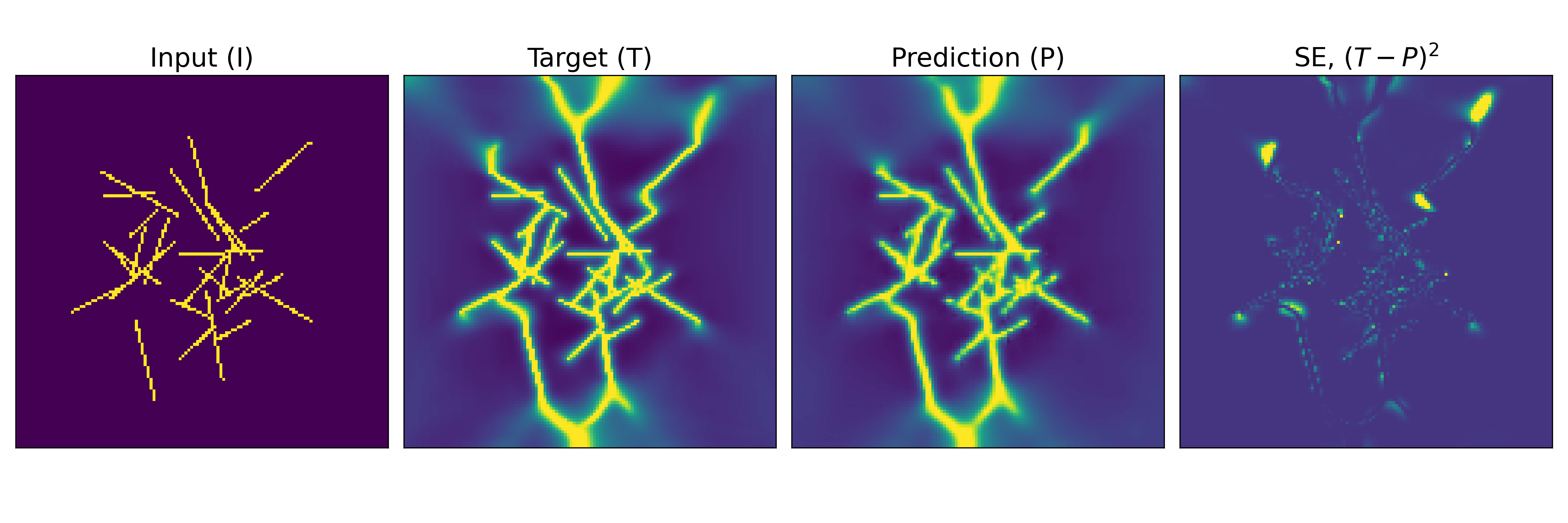}
        \caption{Poseidon-T}
    \end{subfigure}
    \caption{Additional FRAC terminal-state prediction results comparing MORPH-Ti (a) and POSEIDON-T (b) using the same visualization as Fig.~\ref{fig:frac_compare} (input, target, prediction, and squared error $(T-P)^2$).}
    \label{fig:frac_compare_ex}
\end{figure}

\newpage

\begin{figure}[h]
    \centering
    \begin{subfigure}[b]{0.45\linewidth}
        \centering
        \includegraphics[trim={0cm 0cm 0cm 0cm},clip, width=1.0\textwidth]{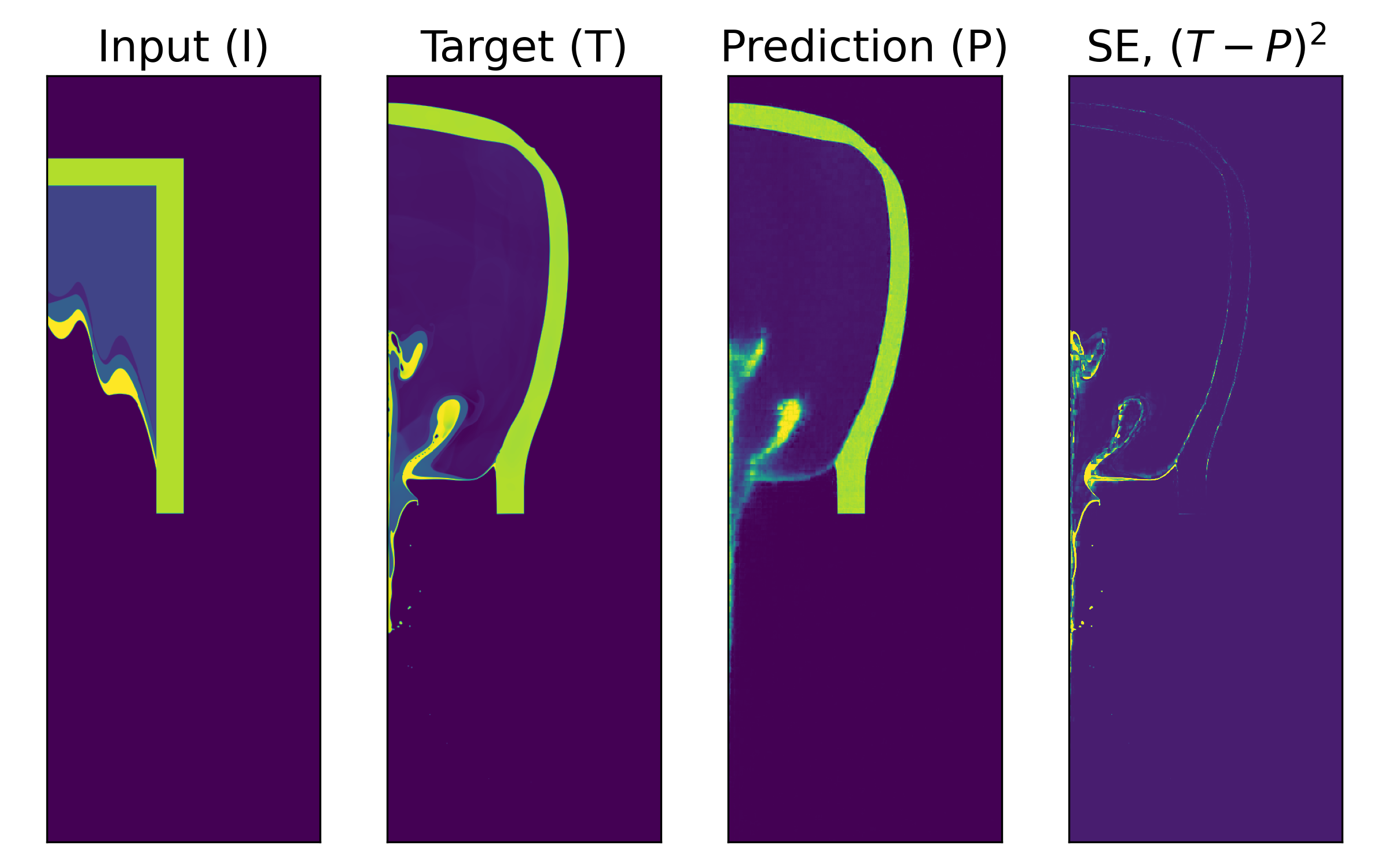}
    \end{subfigure}
    \hspace{1em}
    \begin{subfigure}[b]{0.45\linewidth}
        \centering
        \includegraphics[trim={0cm 0cm 0cm 0cm},clip, width=1.0\textwidth]{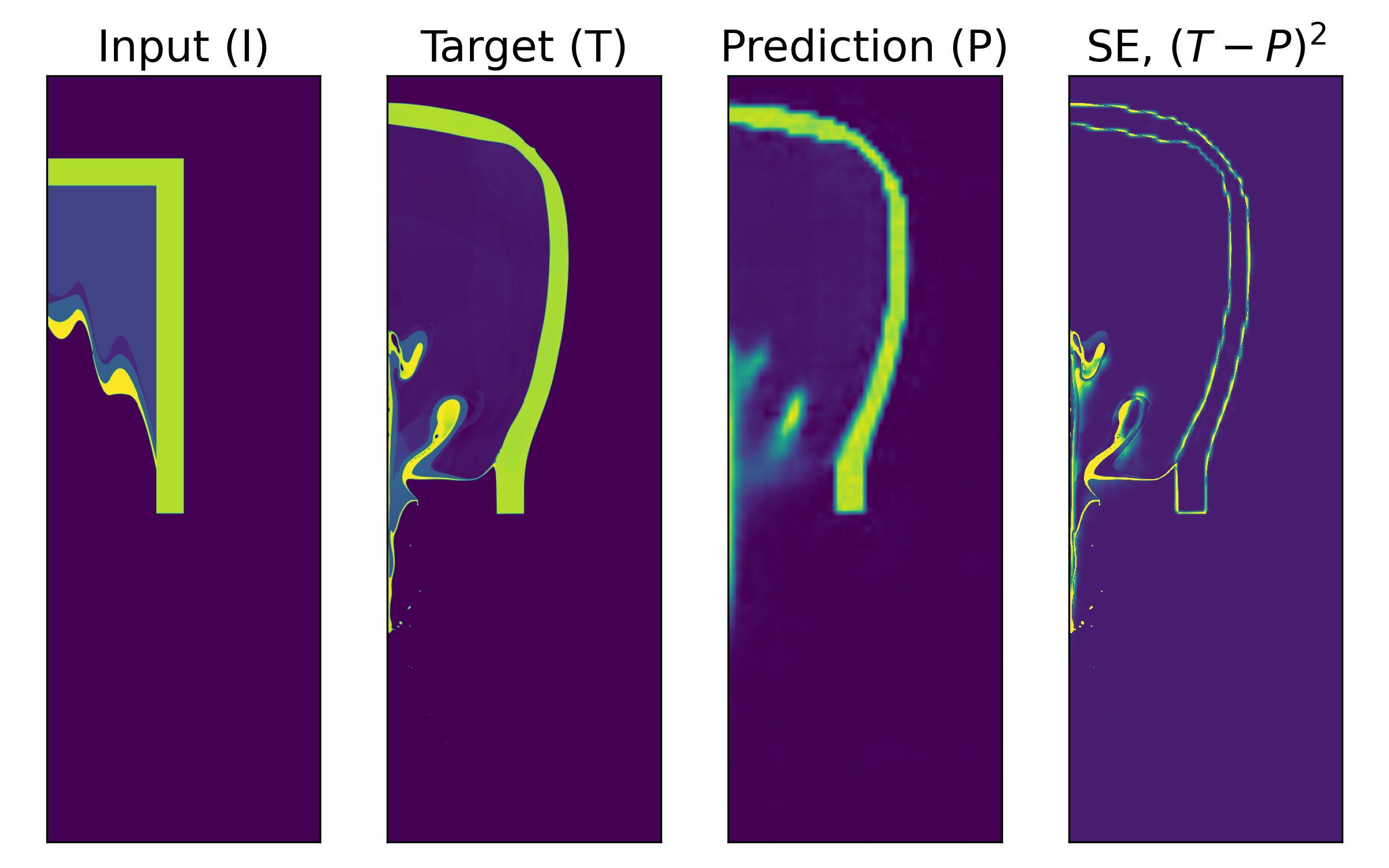}
    \end{subfigure}
    \begin{subfigure}[b]{0.45\linewidth}
        \centering
        \includegraphics[trim={0cm 0cm 0cm 0cm},clip, width=1.0\textwidth]{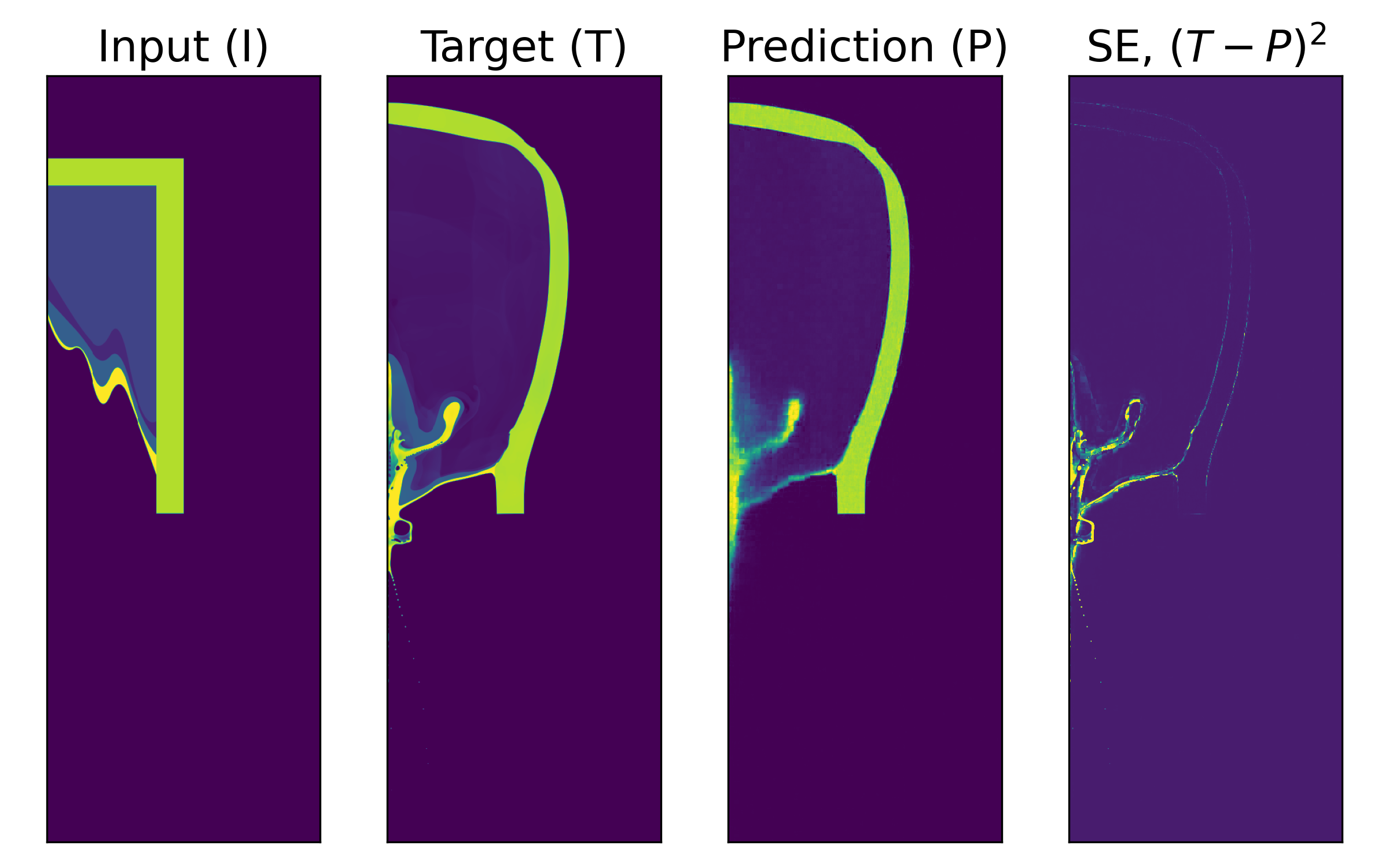}
    \end{subfigure}
    \hspace{1em}
    \begin{subfigure}[b]{0.45\linewidth}
        \centering
        \includegraphics[trim={0cm 0cm 0cm 0cm},clip, width=1.0\textwidth]{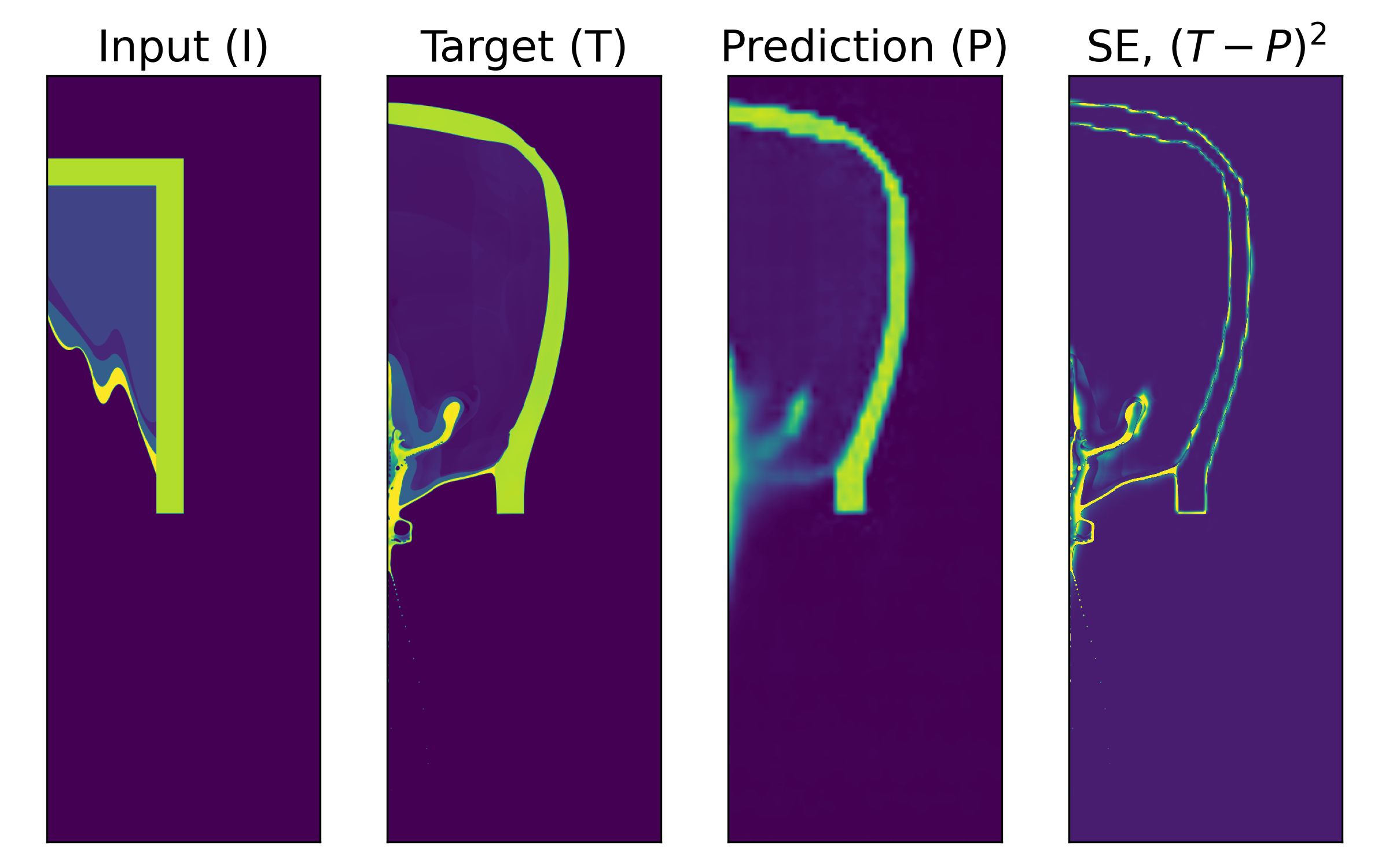}
    \end{subfigure}
    \begin{subfigure}[b]{0.45\linewidth}
        \centering
        \includegraphics[trim={0cm 0cm 0cm 0cm},clip, width=1.0\textwidth]{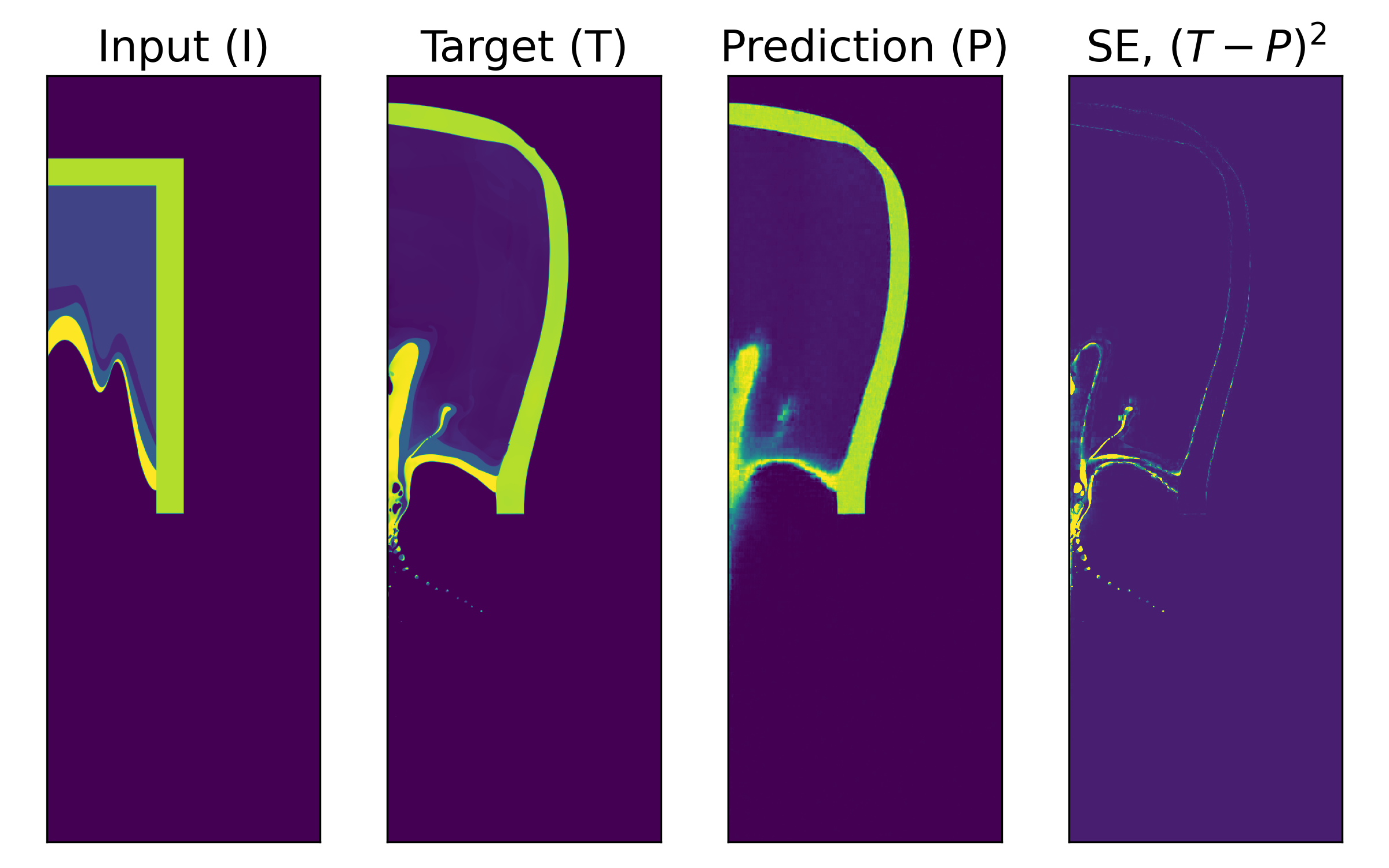}
    \end{subfigure}
    \hspace{1em}
    \begin{subfigure}[b]{0.45\linewidth}
        \centering
        \includegraphics[trim={0cm 0cm 0cm 0cm},clip, width=1.0\textwidth]{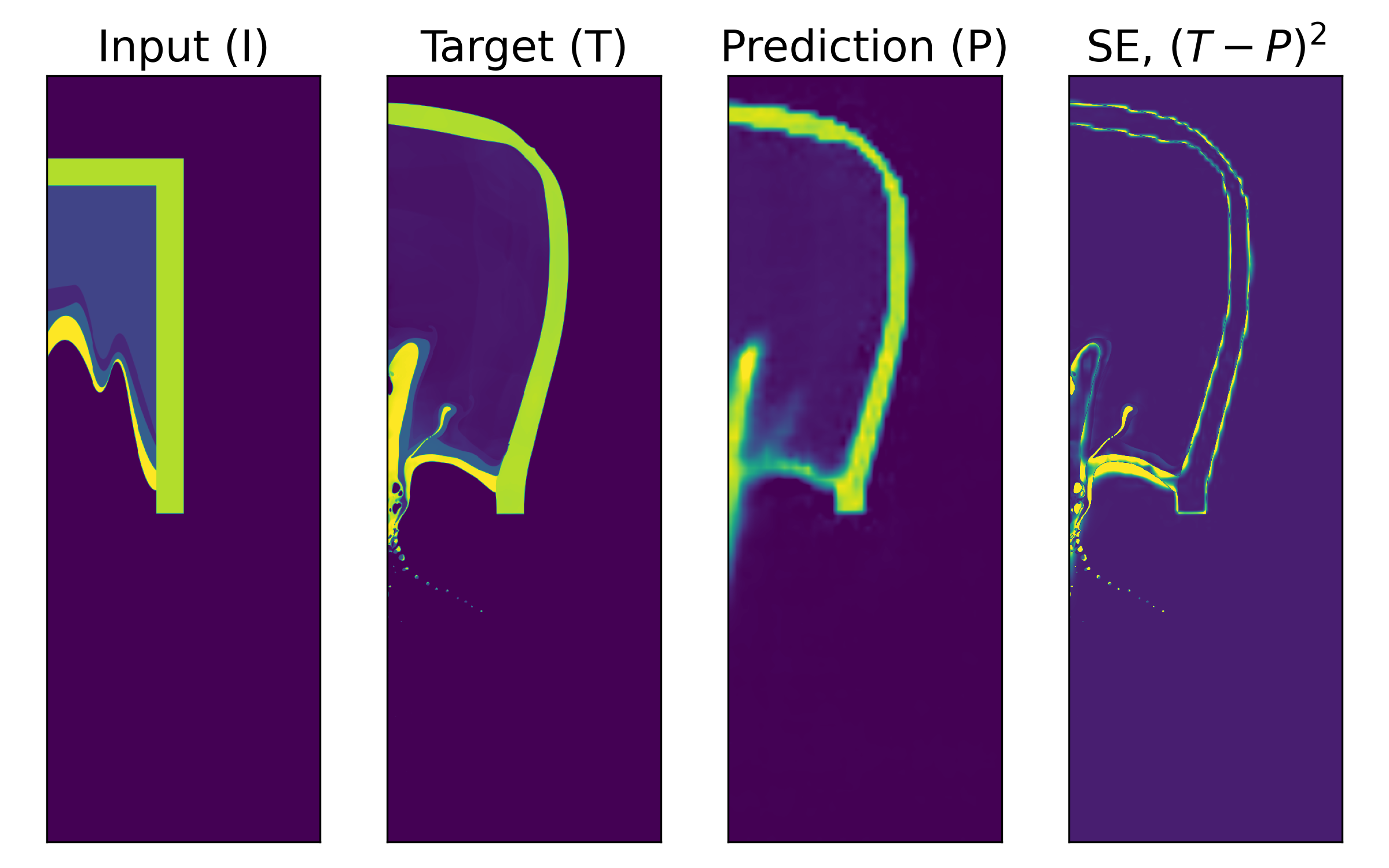}
    \end{subfigure}
    \begin{subfigure}[b]{0.45\linewidth}
        \centering
        \includegraphics[trim={0cm 0cm 0cm 0cm},clip, width=1.0\textwidth]{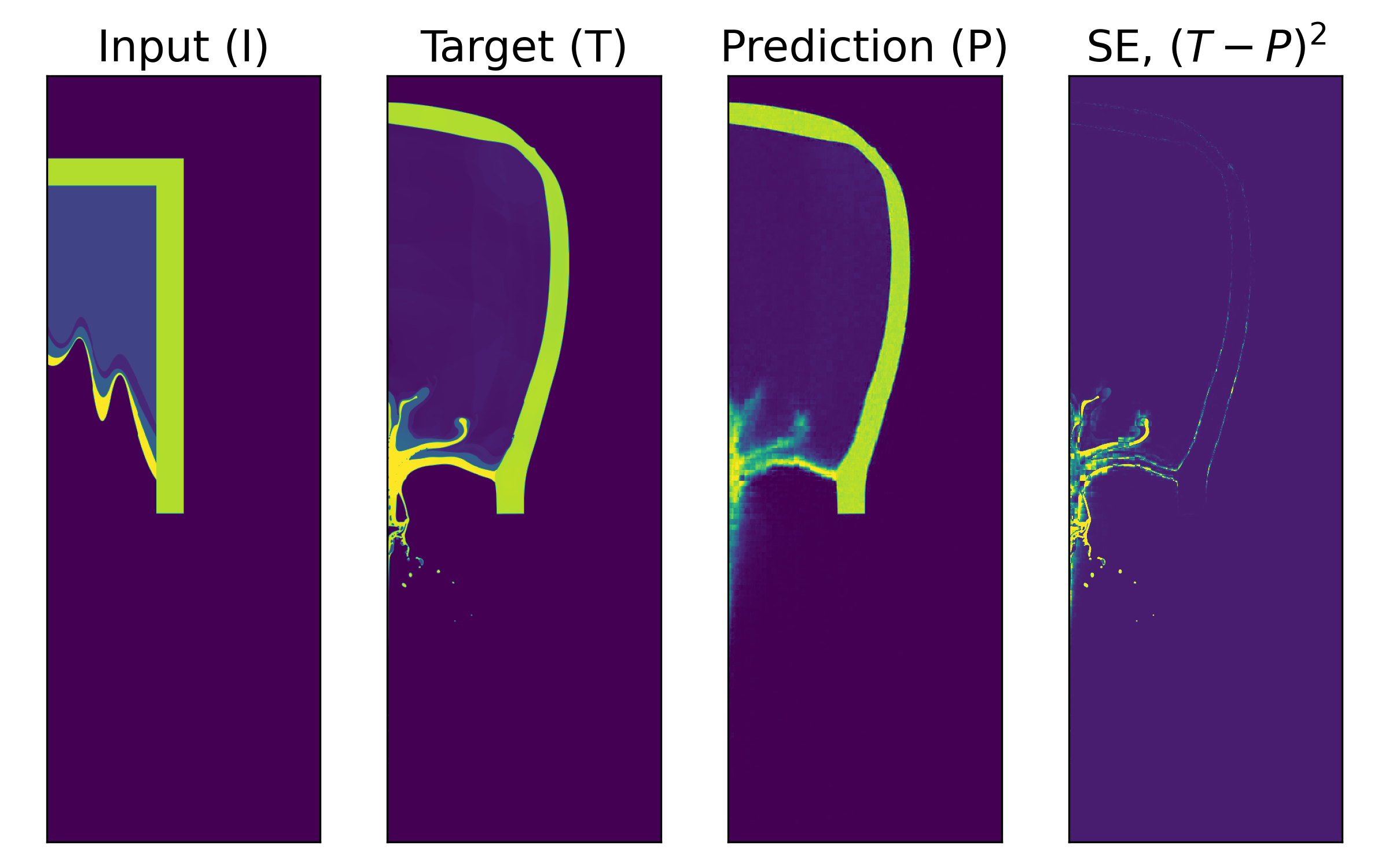}
        \caption{MORPH-Ti}
    \end{subfigure}
    \hspace{1em}
    \begin{subfigure}[b]{0.45\linewidth}
        \centering
        \includegraphics[trim={0cm 0cm 0cm 0cm},clip, width=1.0\textwidth]{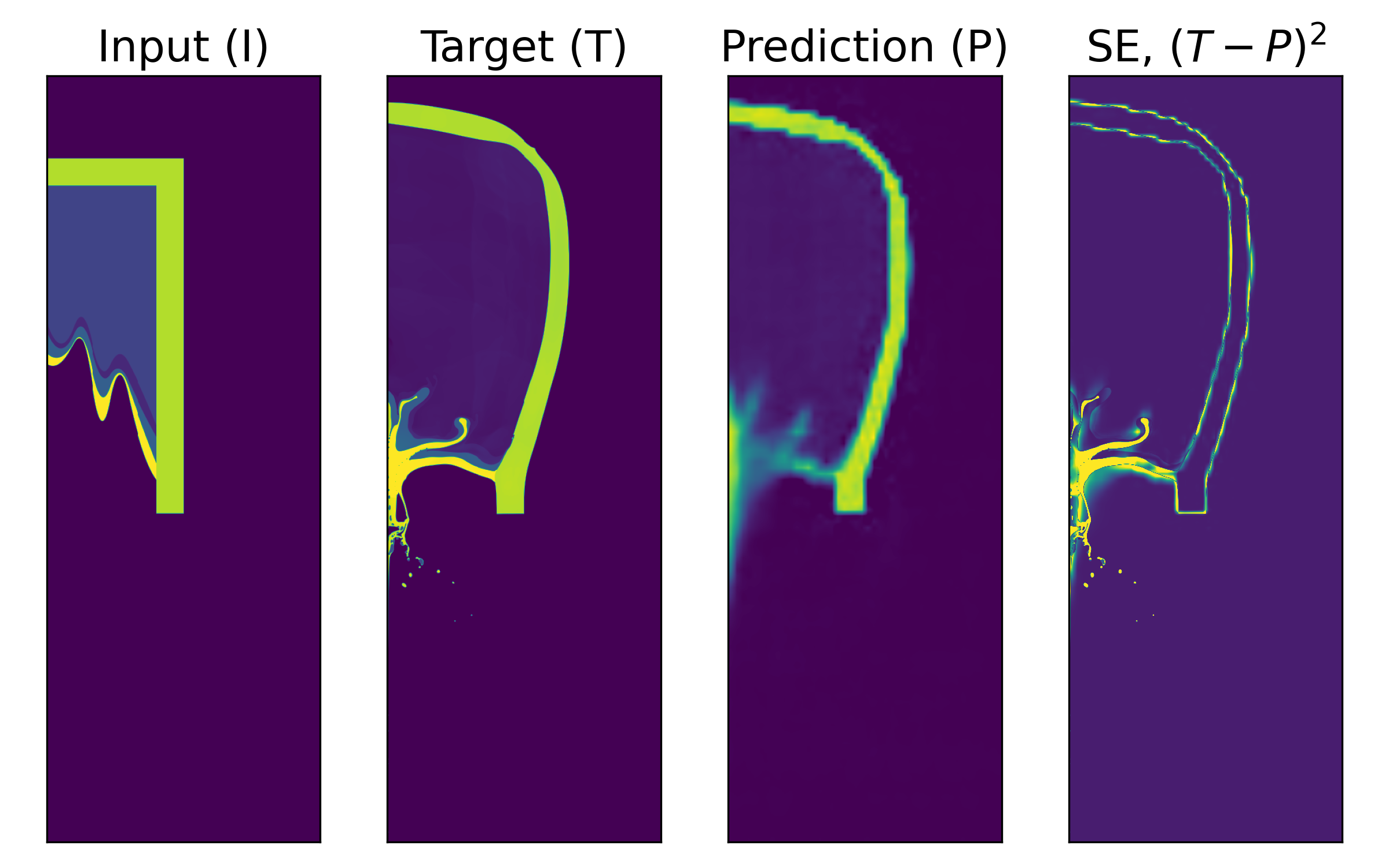}
        \caption{Poseidon-T}
    \end{subfigure}
    \caption{Additional PLI terminal-state prediction results comparing MORPH-Ti (a) and POSEIDON-T (b). Columns show input, target, prediction, and squared error $(T-P)^2$.}
    \label{fig:heat_compare_ex1}
\end{figure}

\newpage

\begin{figure}[h]
    \centering
    \begin{subfigure}[b]{0.45\linewidth}
        \centering
        \includegraphics[trim={0cm 0cm 0cm 0cm},clip, width=1.0\textwidth]{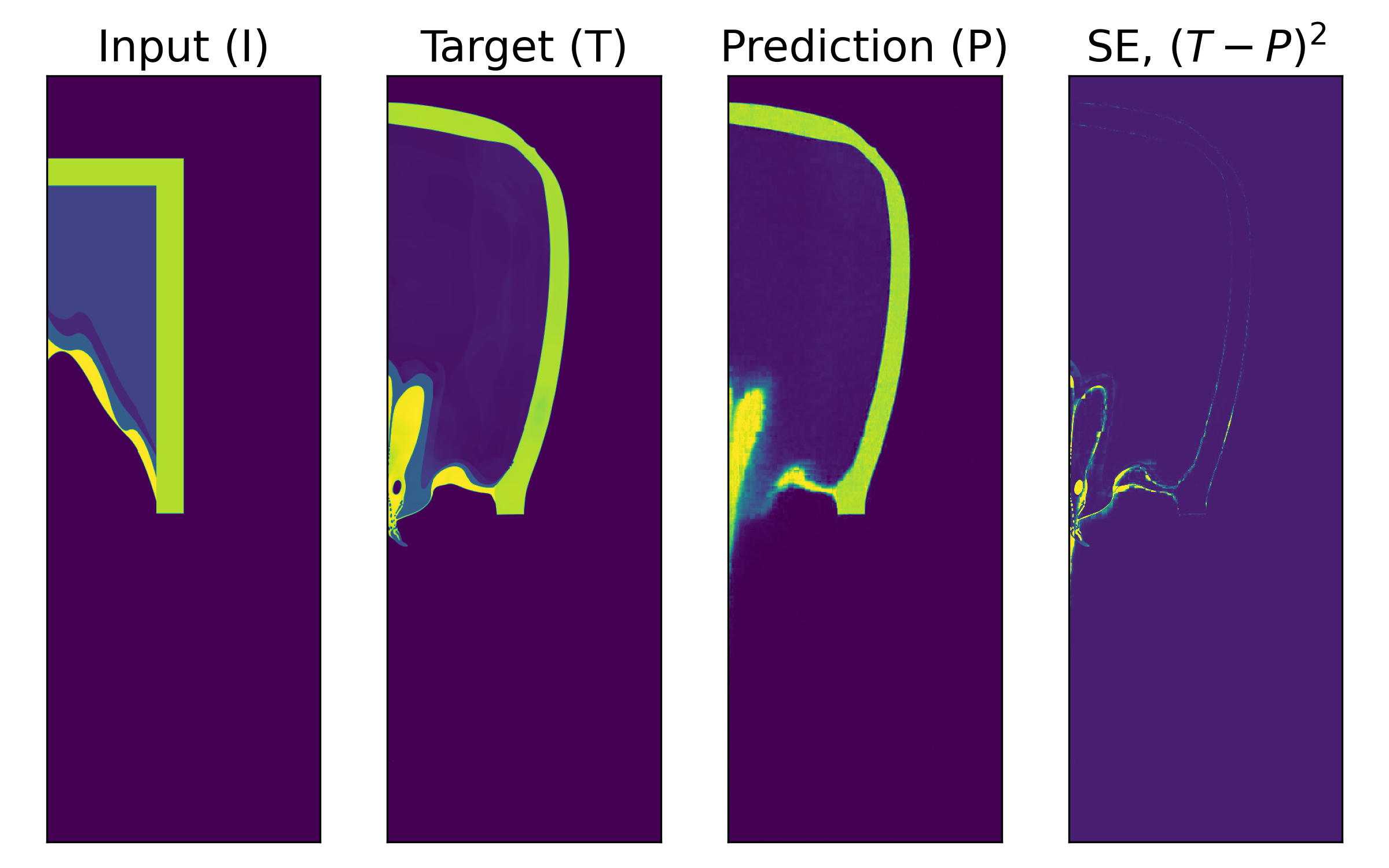}
    \end{subfigure}
    \hspace{1em}
    \begin{subfigure}[b]{0.45\linewidth}
        \centering
        \includegraphics[trim={0cm 0cm 0cm 0cm},clip, width=1.0\textwidth]{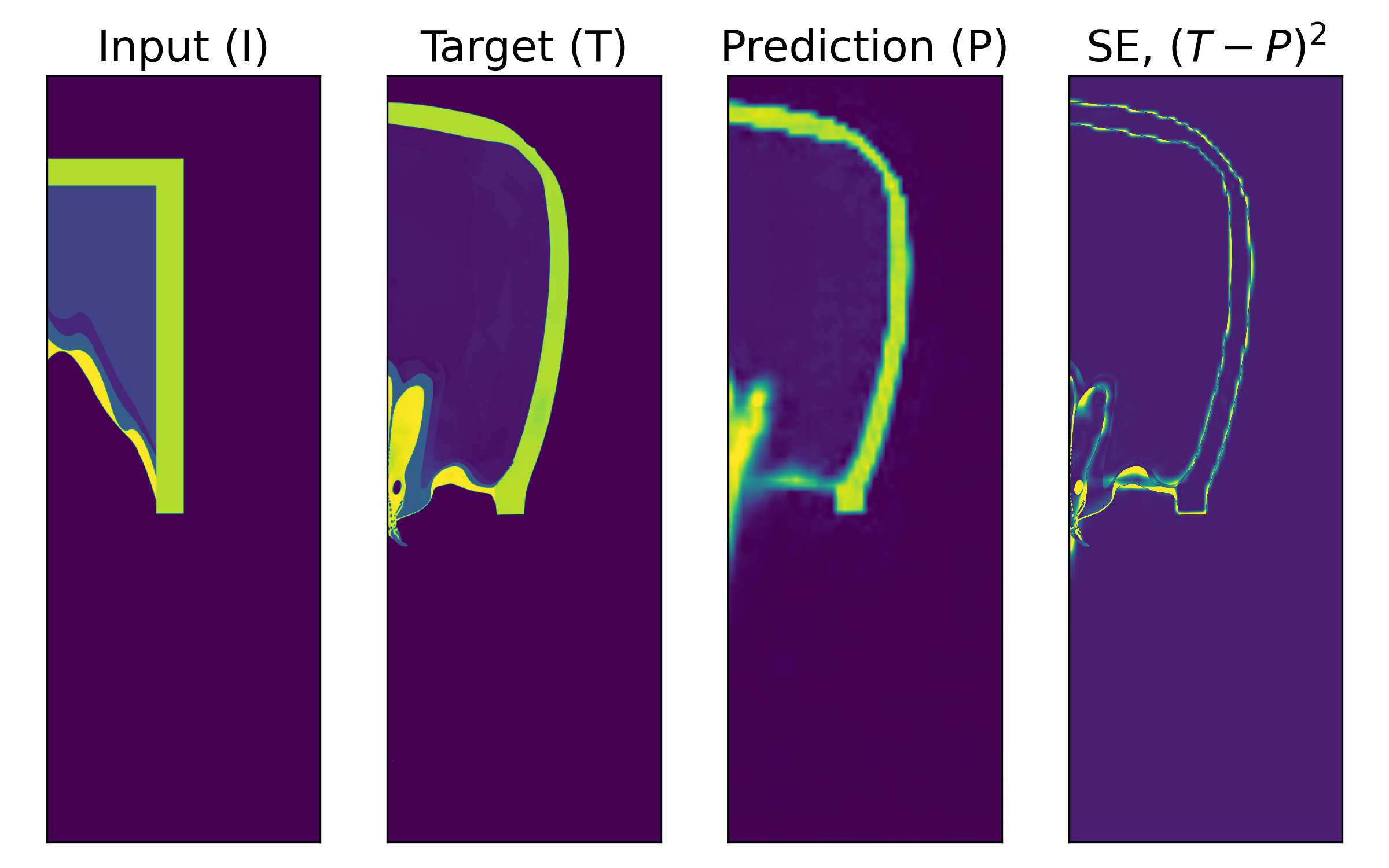}
    \end{subfigure}
    \begin{subfigure}[b]{0.45\linewidth}
        \centering
        \includegraphics[trim={0cm 0cm 0cm 0cm},clip, width=1.0\textwidth]{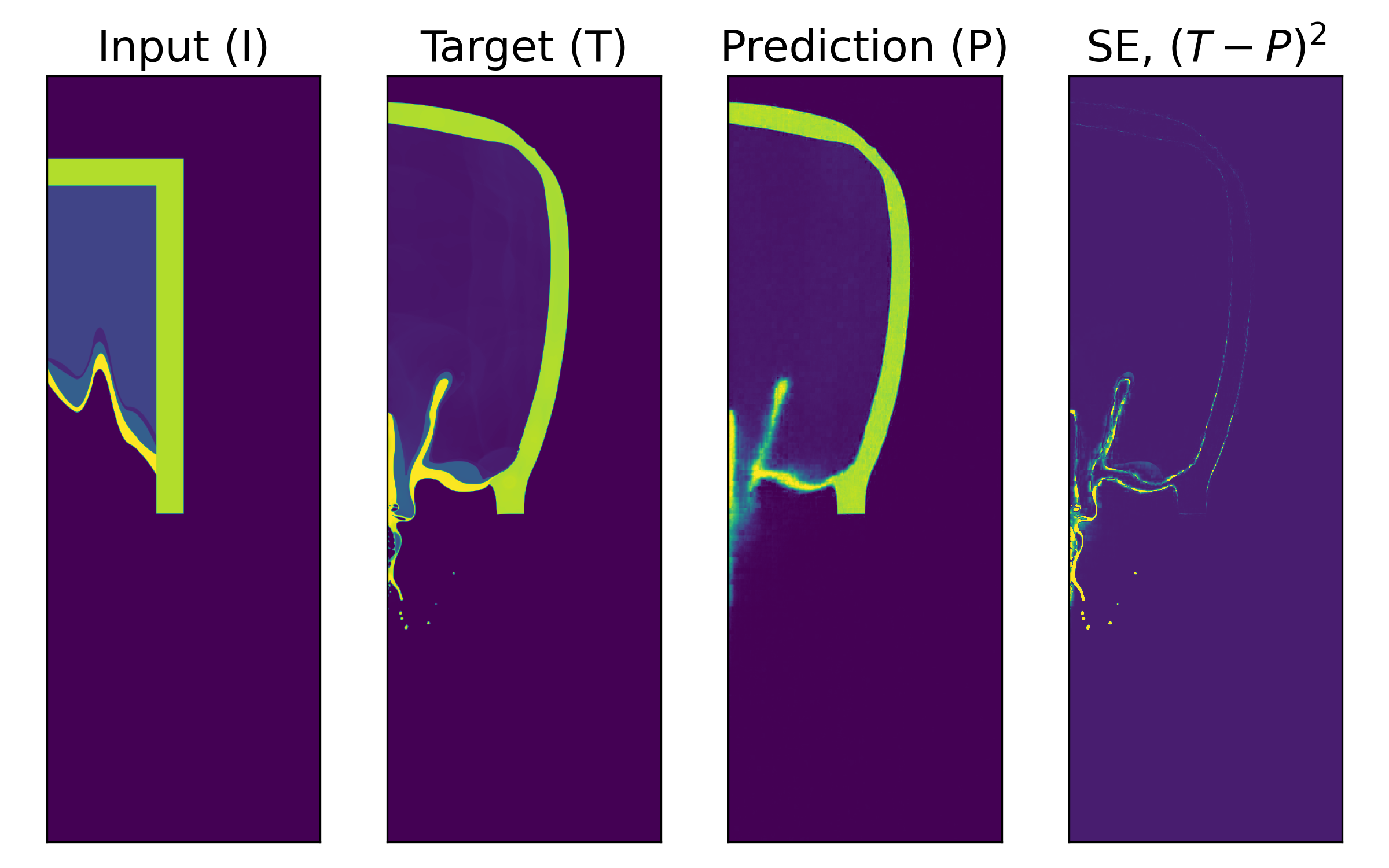}
    \end{subfigure}
    \hspace{1em}
    \begin{subfigure}[b]{0.45\linewidth}
        \centering
        \includegraphics[trim={0cm 0cm 0cm 0cm},clip, width=1.0\textwidth]{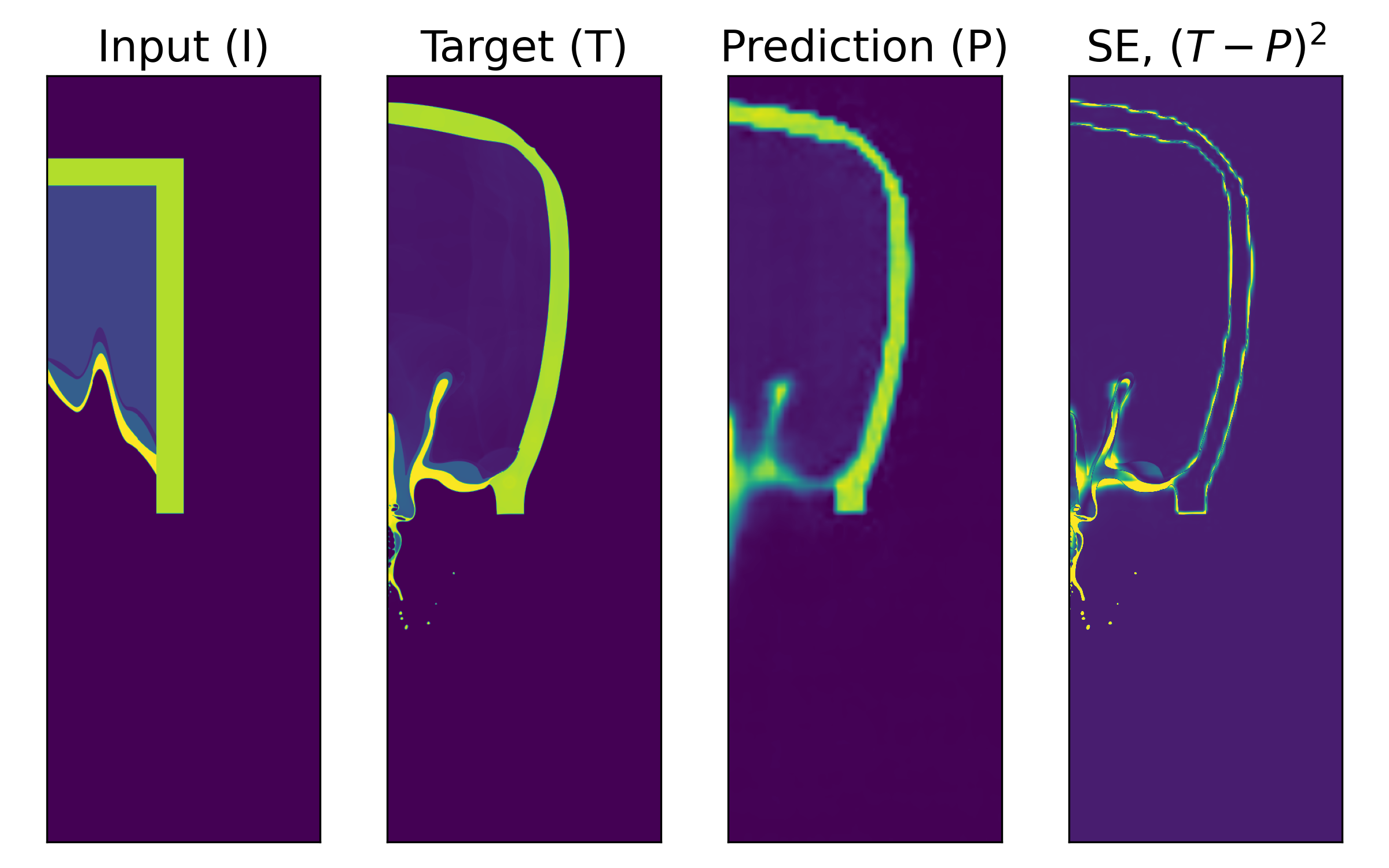}
    \end{subfigure}
    \begin{subfigure}[b]{0.45\linewidth}
        \centering
        \includegraphics[trim={0cm 0cm 0cm 0cm},clip, width=1.0\textwidth]{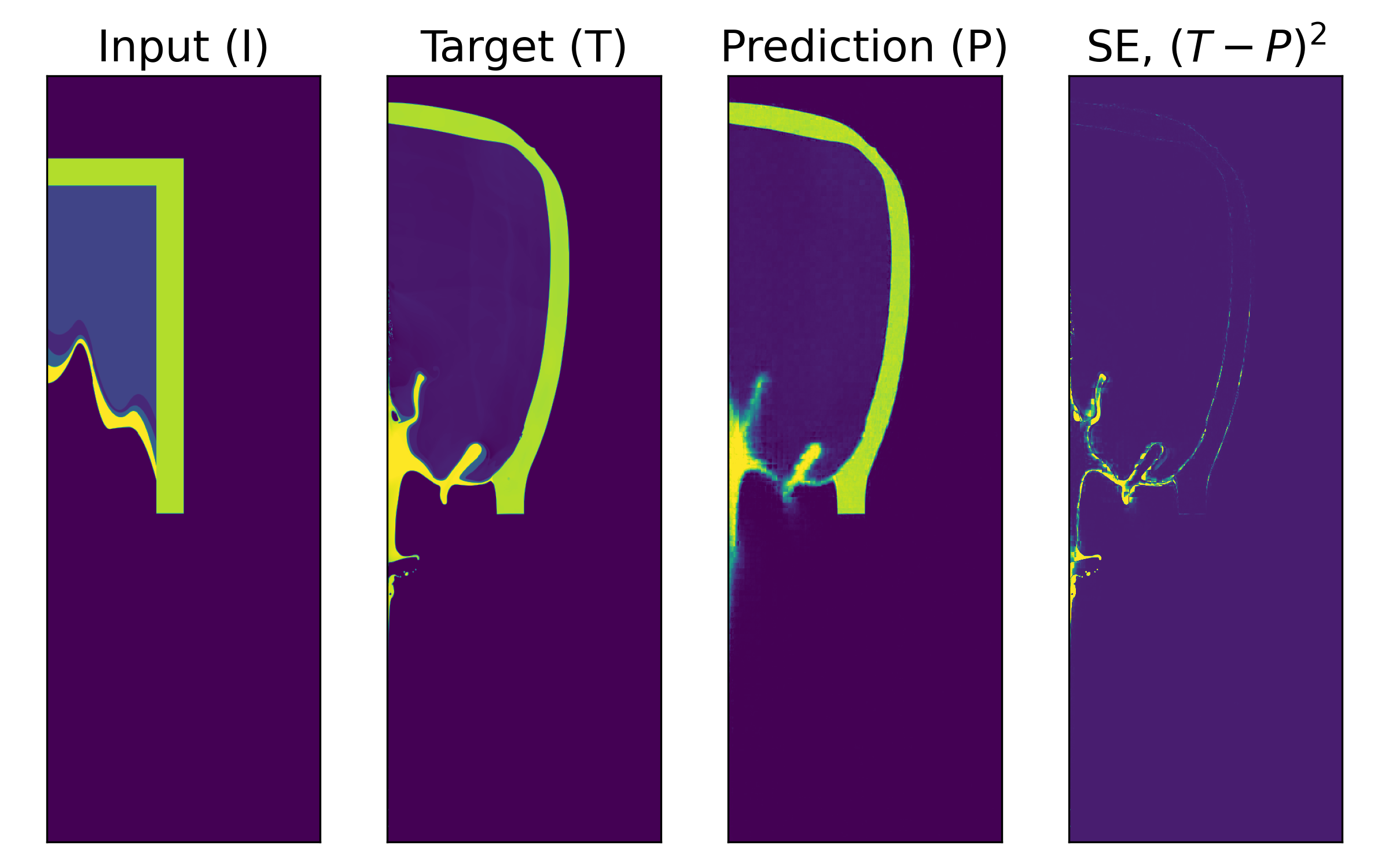}
    \end{subfigure}
    \hspace{1em}
    \begin{subfigure}[b]{0.45\linewidth}
        \centering
        \includegraphics[trim={0cm 0cm 0cm 0cm},clip, width=1.0\textwidth]{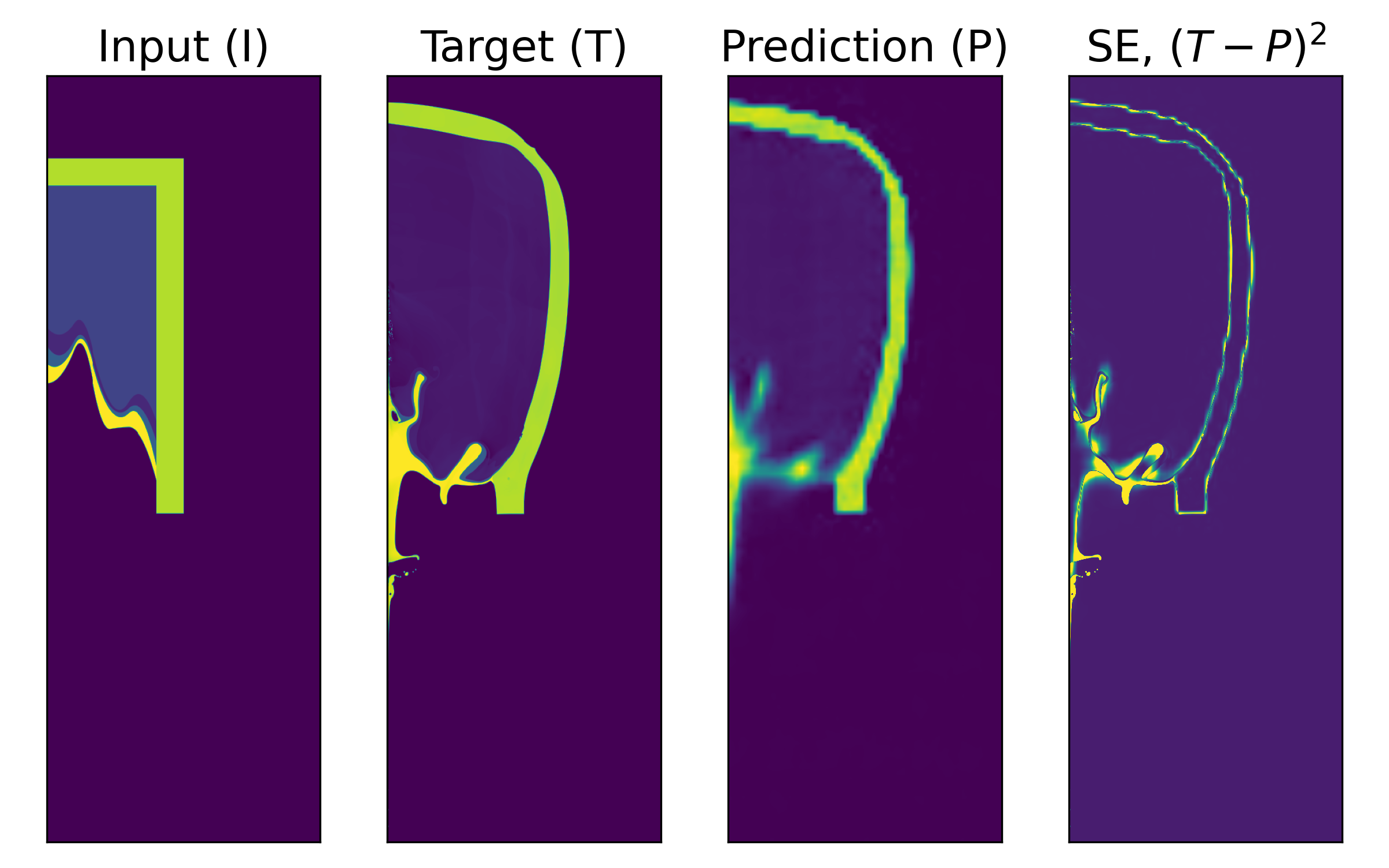}
    \end{subfigure}
    \begin{subfigure}[b]{0.45\linewidth}
        \centering
        \includegraphics[trim={0cm 0cm 0cm 0cm},clip, width=1.0\textwidth]{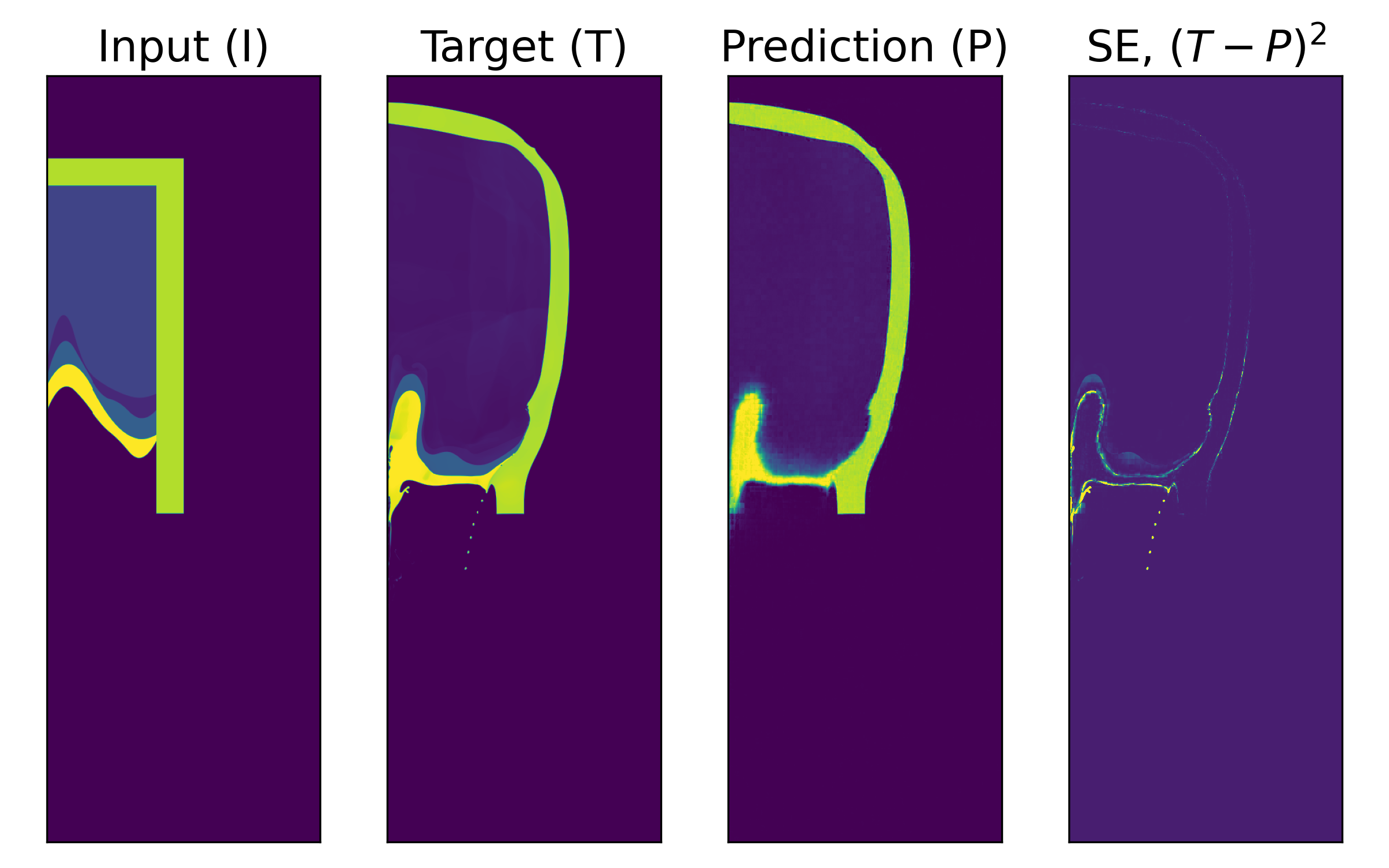}
        \caption{MORPH-Ti}
    \end{subfigure}
    \hspace{1em}
    \begin{subfigure}[b]{0.45\linewidth}
        \centering
        \includegraphics[trim={0cm 0cm 0cm 0cm},clip, width=1.0\textwidth]{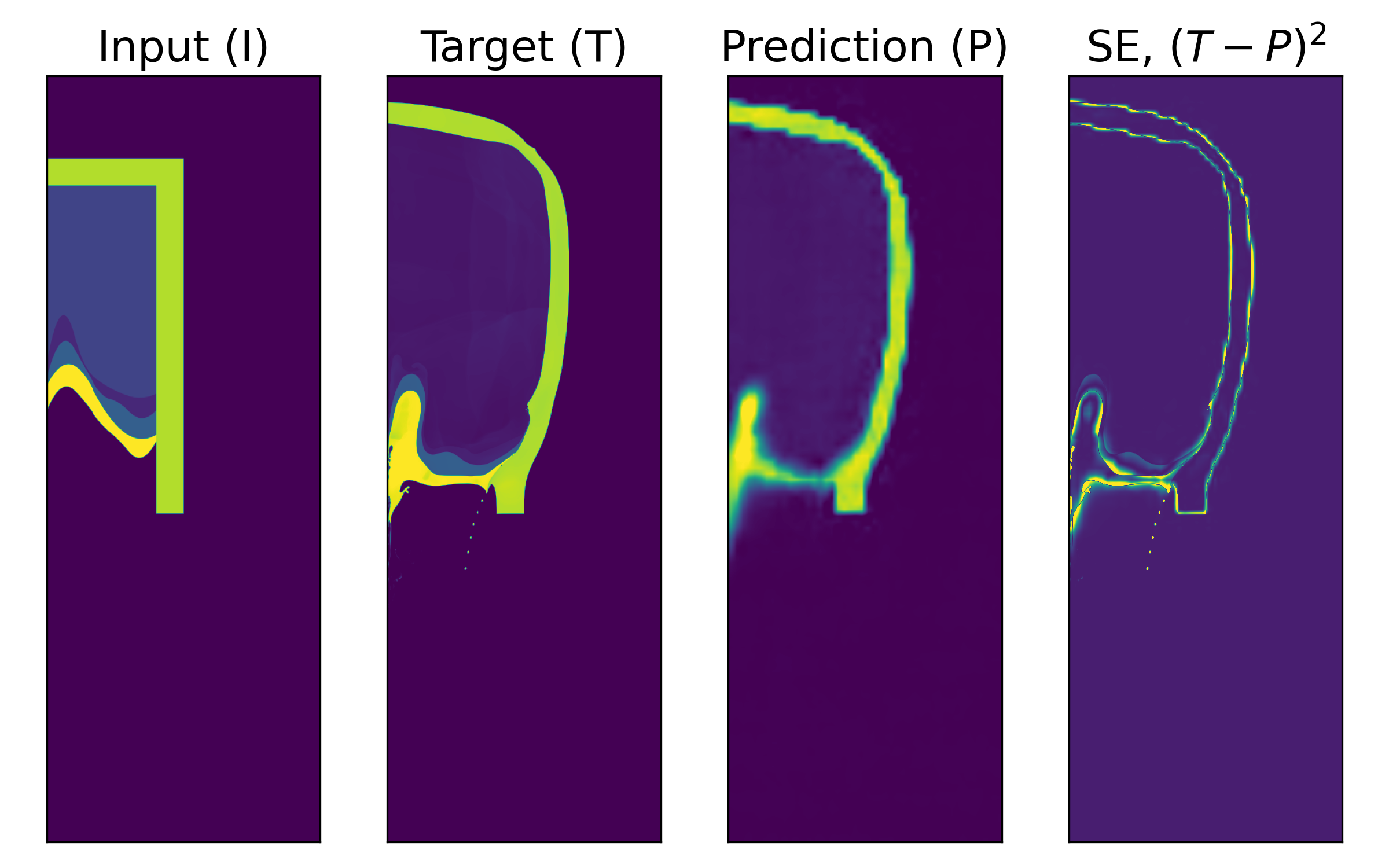}
        \caption{Poseidon-T}
    \end{subfigure}
    \caption{Further PLI terminal-state prediction results comparing MORPH-Ti (a) and POSEIDON-T (b). Columns show input, target, prediction, and squared error $(T-P)^2$.}
    \label{fig:heat_compare_ex2}
\end{figure}

\end{document}

%% file: math_commands.tex
\usepackage{amsmath,amsfonts,bm}

\def\eqref#1{equation~\ref{#1}}

\def\1{\bm{1}}

\DeclareMathAlphabet{\mathsfit}{\encodingdefault}{\sfdefault}{m}{sl}
\SetMathAlphabet{\mathsfit}{bold}{\encodingdefault}{\sfdefault}{bx}{n}

